\documentclass{article}
\usepackage{times}
\usepackage{microtype}        
\usepackage{soul}             
\usepackage{url}              
\usepackage{amsmath, amsfonts, amsthm, amssymb}  
\usepackage{graphicx}         
\usepackage{svg}
\usepackage{array}
\usepackage{cellspace}
\usepackage{multicol}
\usepackage{booktabs}         
\usepackage{makecell}         
\usepackage{multirow}
\usepackage{siunitx}          
\usepackage[table]{xcolor}    
\usepackage{arydshln}         
\usepackage{rotating}         
\usepackage{stfloats}         
\usepackage{verbatim}         
\usepackage[switch]{lineno}
\usepackage{hyperref}
\usepackage{natbib}
\newcommand{\RETURN}{\textbf{return} }
\usepackage[preprint]{icml2026}

\newtheorem{theorem}{Theorem}

\newtheorem{corollary}{Corollary}

\theoremstyle{definition}

\newcounter{problem}

\theoremstyle{remark}
\newtheorem{remark}{Remark}

\icmltitlerunning{Neuronal Attention Circuit (NAC) for Representation Learning}

\begin{document}
\twocolumn[
  \icmltitle{Neuronal Attention Circuit (NAC) for Representation Learning}

  \icmlsetsymbol{equal}{*}

  \begin{icmlauthorlist}
    \icmlauthor{Waleed Razzaq}{yyy}
    \icmlauthor{Yun-Bo Zhao}{yyy,comp}
  \end{icmlauthorlist}

  \icmlaffiliation{yyy}{Department of Automation, University of Science \& Technology of China, Hefei, China}
  \icmlaffiliation{comp}{Institute of Artificial Intelligence, Hefei Comprehensive National Science Center}
  \icmlcorrespondingauthor{Yun-Bo Zhao}{ybzhao@ustc.edu.cn}
  \icmlcorrespondingauthor{Waleed Razzaq}{waleedrazzaq@mail.ustc.edu.cn}
  \icmlkeywords{Continuous-time Attention, Liquid Neural Networks (LNNs)  }
  \vskip 0.15in
]

\printAffiliationsAndNotice{}
\begin{abstract}
Attention improves representation learning over RNNs, but its discrete nature limits continuous-time (CT) modeling. We introduce Neuronal Attention Circuit (NAC), a novel, biologically inspired CT-attention mechanism that reformulates attention logit computation as the solution to a linear first-order ODE with nonlinear interlinked gates derived from repurposing the wiring of \textit{C.\ elegans} Neuronal Circuit Policies (NCPs). NAC replaces dense projections with sparse sensory gates for query-key projections and introduces a sparse backbone network with two heads for computing \textit{content-target} and \textit{learnable time-constant} gates, enabling efficient adaptive dynamics. To improve efficiency and memory consumption, we implement an adaptable, sparse, subquadratic Top-\emph{K} pairwise concatenation mechanism that selectively curates query-key interactions. We provide rigorous theoretical guarantees, including state stability and bounded approximation errors. Empirically, we implement NAC in diverse domains, including irregular time-series, long-range forecasting, lane-keeping for autonomous vehicles, and industrial prognostics. We observe that NAC matches or outperforms in accuracy against several CT state-of-the-art baselines, while being interpretable at the neuron cell level.
\end{abstract}

\section{Introduction}
Learning representations of sequential data in temporal or spatiotemporal domains is essential for capturing patterns and enabling accurate forecasting. Discrete-time Recurrent neural networks (DT-RNNs), such as RNN \cite{rumelhart1985learning,jordan1997serial}, Long-short term memory (LSTM) \cite{hochreiter1997long}, and Gated Recurrent Unit (GRU) \cite{cho2014learning} model sequential dependencies by iteratively updating hidden states to represent or predict future elements in a sequence. While effective for regularly sampled sequences, DT-RNNs face challenges with irregularly sampled data because they assume uniform time intervals. In addition, vanishing gradients can make it difficult to capture long-term dependencies \cite{hochreiter1998vanishing}. \\
Continuous-time RNNs (CT-RNNs) \cite{rubanova2019latent} model hidden states as ordinary differential equations (ODEs), allowing them to process inputs that arrive at arbitrary or irregular time intervals. Mixed-memory RNNs (mmRNNs) \cite{lechner2022mixed} build on this idea by separating memory compartments from time-continuous states, helping maintain stable error propagation while capturing continuous-time dynamics. Liquid neural networks (LNNs) \cite{hasani2021liquid,hasani2022closed} take a biologically inspired approach by assigning variable time-constants to hidden states, improving adaptability and robustness, although vanishing gradients can still pose challenges during training. \\
The scaled-dot-product-attention (SDPA) mechanism \cite{vaswani2017attention} mitigate this limitation by treating all time steps equally and allowing models to focus on the most relevant observations. It computes the similarity between queries (\(q\)) and keys (\(k\)), scaling by the key dimension to keep the gradients stable. Multihead Attention (MHA) \cite{vaswani2017attention} extends this by allowing the model to attend to different representation subspaces in parallel. Variants such as Sparse Attention \cite{zhang2021sparse}, BigBird \cite{zaheer2020big}, and Longformer \cite{beltagy2020longformer} modify the attention pattern to reduce computational cost, particularly for long sequences, by attending only to selected positions rather than all pairs. Even with these improvements, the computation paths in these methods are discrete in nature, limiting their ability to model continuous trajectories often captured by their CT counterparts. \\
Recent work has explored bridging this gap through the Neural-ODE \cite{chen2018neural} formulation. mTAN \cite{shukla2021multi} learns CT embeddings and uses time-based attention to interpolate irregular observations into a fixed-length representation for downstream encoder-decoder modeling. Continuous-time Attention (CTA) \cite{chien2021continuous} embeds a CT-attention mechanism within a Neural ODE, allowing attention weights and hidden states to evolve jointly over time. Nevertheless, it remains computationally intensive and sensitive to the accuracy of the ODE solver. ODEFormer \cite{d2023odeformer} trains a sequence-to-sequence transformer on synthetic trajectories to infer symbolic ODE systems directly from noisy, irregular data, although it struggles with chaotic systems and generalization beyond observed conditions. ContiFormer \cite{chen2023ContiFormer} builds a CT-Transformer by pairing ODE-defined latent trajectories with a time-aware attention mechanism to model dynamic relationships in data. \\
Despite these advancements, a persistent and underexplored gap remains in developing a biologically inspired attention mechanism that seamlessly integrates CT dynamics with the abstraction of the brain's connectome. Building on this, we propose a novel attention mechanism called the \textit{Neuronal Attention Circuit} (NAC), in which attention logits are computed as the solution to a first-order ODE modulated by nonlinear, interlinked gates derived from repurposing Neuronal Circuit Policies (NCPs) from the nervous system of \textit{C.elegans} nematode (refer to Appendix~\ref{appendix:NCPs} for more information). Unlike SDPA, which projects query, key, value vectors through a linear layer, NAC employs a sensory gate to transform input features and a backbone network to model nonlinear interactions, with multiple heads producing outputs structured for attention logit computation. Based on the solutions to ODE, we define three computation modes: (i) Exact, using the closed-form ODE solution; (ii) Euler, approximating the solution via \textit{explicit Euler} integration; and (iii) Steady, using only the steady-state solution, analogous to SDPA scores. To reduce computational complexity, we implement a sparse subquadratic Top-\emph{K} pairwise concatenation algorithm that selectively curates query-key inputs. We evaluate NAC across multiple domains, including irregularly sampled time-series, long-range forecasting, autonomous vehicle lane-keeping, and Industry 4.0, and compare it to state-of-the-art baselines. NAC consistently matches or outperforms these models, while runtime and peak memory benchmarks place it between CT-RNNs in terms of speed and CT-attentions in terms of memory requirements. We also highlight the interpretability of NAC at the cell level.

Our contributions are summarized as:
\begin{itemize}
    \item We proposed Neuronal Attention Circuit (NAC), a novel biologically-inspired attention mechanism that reformulates attention logits through an input-dependent linear ODE modulated with nonlinear interlinked gates.
    \item We repurposed \textit{C.elegans} nematode's  Neuronal Circuit Policies (NCPs) wiring mechanism to derive the nonlinear gating mechanism to introduce sparsity and cell-level interpretability into the attention mechanism.
    \item We evaluate NAC in a broad set of tasks including (i) irregular time-series; (ii) long-range dependencies; (iii) autonomous-vehicle lane-keeping control; and (iv) Industry~4.0.
\end{itemize}
The rest of the paper is organized as follows. Section~\ref{sec:related} reviews the related literature and highlights the contributions. Section~\ref{sec:NAC} presents the comprehensive architectural details of proposed algorithm. Section~\ref{sec:eval} comprehensively evaluate the proposed algorithm in a broad set of learning tasks. Section~\ref{sec:discuss} discuss the limitations and possible future directions. Section~\ref{sec:conclude} conclude the paper.

\section{Related Works}\label{sec:related}
\textbf{Bio-inspired models:} LNNs~\cite{hasani2021liquid, hasani2022closed} introduce input-dependent hidden states dynamics by allowing the integration time constants of recurrent hidden states to vary through nonlinear gating mechanisms. These models are often implemented within the NCPs~\cite{lechner2018neuronal, lechner2020neural} framework, leveraging the sparse connectivity of the \textit{C.elegans} connectome to produce neural dynamics. NAC draws on these input-dependent mechanisms but applies them to attention score computation rather than hidden state evolution.

\textbf{CT Attention:} Prior work incorporates attention mechanisms into Neural ODE frameworks, allowing hidden states and attention weights to evolve jointly over time~\cite{chien2021continuous}, couples latent dynamics with time-aware attention~\cite{chen2023ContiFormer}, trains transformers to learn underlying CT sequence representations~\cite{d2023odeformer}, formulates entire transformer blocks as a single ODE with optimal-transport regularization~\cite{kan2025ot}, or evolves the full attention matrix over pseudo-time via PDEs~\cite{zhang2025continuous}. However, the underlying attention score computation path in all these approaches remains inherently discrete. NAC rethinks attention score computation as the solution to a first-order ODE with an input-dependent nonlinear gate, turning the discrete attention itself into an adaptive continuous-depth process.

\section{Neuronal Attention Circuit (NAC)}\label{sec:NAC}
We propose a simple alternative formulation of the attention logits $a$ (refer to Appendix~\ref {appendix:attention} for more information), interpreting them as the solution to a first-order linear ODE modulated by nonlinear, interlinked gates:
\begin{equation}
\frac{da_t}{dt} = -\,\underbrace{f_{\omega_\tau}([\mathbf{q};\mathbf{k}],t,\theta_{\omega_\tau})}_{\omega_\tau(\mathbf{u})} a_t + \underbrace{f_\phi([\mathbf{q};\mathbf{k}],t,\theta_{\phi})}_{\phi(\mathbf{u})},
\label{eq:lq_diff}
\end{equation}
where $\mathbf{u} = [\mathbf{q}; \mathbf{k}]$ denotes the sparse Top-\emph{K} concatenated query--key input. $\omega_\tau$ represents a \textit{learnable time-constant} gate head, and $\phi$ denotes a nonlinear \textit{content-target} head. Both gates are parameterized by a backbone network derived from repurposing NCPs. We refer to this formulation as the Neuronal Attention Circuit (NAC). It enables the logit $a_t$ to evolve dynamically with input-dependent, variable time constants, mirroring the adaptive temporal dynamics found in \textit{C.elegans} nervous systems while improving computational efficiency and expressiveness. Moreover, it introduces continuous depth into the attention mechanism, bridging discrete-layer computation with dynamic temporal evolution.\\
\textbf{Motivation behind this formulation:} The proposed formulation is loosely motivated by the input-dependent time-constant mechanism of LNNs, a class of CT-RNNs inspired by biological nervous systems and synaptic transmission. In LNNs, the dynamics of non-spiking neurons are described by a linear ODE with nonlinear, interlinked gates: \(\frac{d\mathbf{x_t}}{dt} =- \frac{\mathbf{x_t}}{\tau} + \mathbf{S_t}, \) where $\mathbf{x_t}$ denotes the hidden state and $\mathbf{S_t} \in \mathbb{R}^M$ represents nonlinear synapse defined as $f(\mathbf{x_t}, \mathbf{u}, t, \theta)(A - \mathbf{x_t})$. Here, $A$ and $\theta$ are learnable parameters. Plugging $\mathbf{S_t}$ yields \(
\frac{d\mathbf{x_t}}{dt} =-\left[\frac{1}{\tau} + f(\mathbf{x_t}, \mathbf{u}, t, \theta)\right] \mathbf{x_t} + f(\mathbf{x_t}, \mathbf{u}, t, \theta)A\). LNNs are known for their strong expressivity, stability, and performance in irregularly sampled time-series modeling \cite{hasani2021liquid,hasani2022closed}.\\
\textbf{NAC's forward-pass update using ODE solver:} The state of NAC at time $t$ can be computed using a numerical ODE solver that simulates the dynamics from an initial state $a_0$ to $a_t$. The solver discretizes the continuous interval $[0, T]$ into steps $[t_0, t_1, t_2, \dots, t_n]$, with each step updating the state from $t_i$ to $t_{i+1}$. For our purposes, we use the \textit{explicit Euler} solver, which is simple, efficient, and easy to implement. Although methods such as Runge-Kutta may offer higher accuracy, their computational overhead makes them less suitable for large-scale neural simulations that require numerous updates, especially since the logits are normalized, and exact precision is not necessary. Let the step size be $\Delta t$, with discrete times $t_n = n \Delta t$ and logit states $a_n = a(t_n)$. Using the \textit{explicit Euler} method, the update is\\
\begin{equation}
a_{n+1} = a_n + \Delta t \,(-\omega_\tau a_n + \phi).
\label{eq:euler}
\end{equation}
\textbf{Closed-form (Exact) Computation of NAC:} We derive the analytical solution for Eqn.~\ref{eq:lq_diff}. The gates $\phi$ and $\omega_{\tau}$ are time-varying and depend on the input sequence at each time step. However, to enable an efficient closed-form forward-pass update, we apply a piecewise constant assumption (locally frozen-coefficient approximation~\cite{john1952integration}) over a small interval. Under this assumption, with initial condition $a_0$, the analytical closed-form solution is as follows:
\begin{equation}
a_t
= \underbrace{a^*}_{\text{steady-state}}
+ \underbrace{(a_0 - a^*) \, e^{-\omega_\tau t}}_{\text{transient}}
\label{eq:lq_exact}
\end{equation}
Here, \(a^* = \phi/\omega_\tau\) is the steady-state solution. The detailed derivation and analysis are provided in Appendix \ref{appendix:analyzing_exact}.

\subsection{Stability Analysis of NAC}\label{sec:stability_analysis}
We now investigate the stability bounds of NAC under both the ODE-based and the Closed-Form formulations.

\subsubsection{State Stability}
We analyze state stability in both single-connection and multi-connection settings. This analysis establishes the boundness of the attention logit state trajectory, ensuring that, under positive decay rates, the dynamics remain well-behaved without divergence or overshoot.
\begin{theorem}[State Stability]\label{theorem:state_stability}
Let $a_t^{(i)}$ denote the state of the $i$-th attention logit governed by \(da_t^{(i)}/dt = -\omega_\tau a_t^{(i)} + \phi\). Assume that $\phi$ and $\omega_\tau$ decompose across $M$ incoming connections as \(\phi = \sum_{j=1}^M f_\phi([\mathbf{q}_i; \mathbf{k}_j])\), and \(\omega_\tau = \sum_{j=1}^M f_{\omega_\tau}([\mathbf{q}_i; \mathbf{k}_j]),\) with $f_{\omega_\tau}> 0$. Define the per-connection equilibrium $A_{i,j} = f_\phi([\mathbf{q}_i; \mathbf{k}_j]) / f_{\omega_\tau}([\mathbf{q}_i; \mathbf{k}_j])$, and let $A_i^{\min} = \min_j A_{i,j}$ and $A_i^{\max} = \max_j A_{i,j}$. Then for any finite horizon $t \in [0,T]$, the state trajectory satisfies
\begin{equation}\min(0, A_i^{\min}) \;\leq\; a_t^{(i)} \;\leq\; \max(0, A_i^{\max}),\end{equation}
provided the initial condition $a_i(0)$ lies within this range. In the special case of a single connection ($M=1$), the bounds collapse to
\begin{equation}
\min(0, A_i) \;\leq\; a_t^{(i)} \;\leq\; \max(0, A_i),
\end{equation}
where $A_i = f_\phi/ \omega_\tau$ is exactly the steady-state solution from Eqn.~\ref{eq:lq_exact}. The proof is provided in the Appendix \ref{appendix:theorem_state}.
\end{theorem}

\subsubsection{Closed-form Error \& Exponential Bounds}
We now examine the asymptotic stability, error characterization, and exponential boundedness of the closed-form formulation. We begin by quantifying the deviation of the trajectory from its steady-state solution. Define the instantaneous error
\begin{equation}
\varepsilon_t = a_t - a^*,
\end{equation}
which measures the distance of the system state to equilibrium at time \(t\). From Eqn.~\ref{eq:lq_exact}, the error admits the exact representation
\begin{equation}
\varepsilon_t = (a_0-a^*) e^{-\omega_\tau t}
\end{equation}
In particular, the pointwise absolute error is given by
\begin{equation}
|\varepsilon_t| = |a_0-a^*|\, e^{-\omega_\tau t}
\end{equation}
This reveals that convergence is not merely asymptotic but follows an exact exponential law, controlled by the rate parameter \(\omega_\tau\). This yields the following finite-time guarantee.
\begin{corollary}[Exponential decay bound]
If \(\omega_\tau>0\), then for all \(t\ge 0\),\\
\begin{equation}
|a_t-a^*| \le |a_0-a^*|\, e^{-\omega_\tau t}.
\end{equation}
\end{corollary}
\emph{Remark:} If \(\omega_\tau>0\) then \(\lim_{t\to\infty} e^{-\omega_\tau t}=0\), therefore \(a_t\to a^*\). The convergence is exponential with rate \(\omega_\tau\). If \(\omega_\tau<0\) then \(e^{-\omega_\tau t}=e^{|\omega_\tau|t}\) diverges so that \(a_t\) grows exponentially away from \(a^*\) in magnitude (unless initial offset \(a_0-a^*=0\), a measure-zero case). If \(\omega_\tau=0\) the ODE is \(\dot a=\phi\) and the solution is linear in \(t\) (unless \(\phi=0\)). For bounded dynamics that converge to an interpretable steady-state, it is required that \(\omega_\tau>0\).\\
\begin{corollary}[Uniform initialization]
If the initialization is known only to belong to a bounded set, i.e.,
\(|a_0-a^*|\le M\) for some \(M>0\), then the error admits the uniform bound
\begin{equation}
|a_t-a^*| \le M e^{-\omega_\tau t}.
\end{equation}
\end{corollary}
\emph{Remark:} This bound highlights that exponential convergence holds uniformly across all admissible initial conditions, with the constant \(M\) capturing the worst-case deviation.\\
\begin{corollary}[Convergence time to \(\delta\)-accuracy]
 A natural operational question is the time required to achieve a target tolerance \(\delta>0\). Solving
\begin{equation}
|a_0-a^*| e^{-\omega_\tau t} \le \delta,
\end{equation}
We obtain the threshold
\begin{equation}
t \ge \frac{1}{\omega_\tau}\ln\frac{|a_0-a^*|}{\delta}.
\end{equation}
\end{corollary}
\emph{Remark:} The convergence time rate is inversely proportional to \(\omega_\tau\), and the required time scales only logarithmically at the accuracy level \(1/\delta\). Intuitively, a larger \(\omega_\tau\) accelerates contraction toward equilibrium, yielding faster attainment of any prescribed tolerance.

\begin{figure*}[ht!]
\centering
\includegraphics[width=1.0\textwidth]{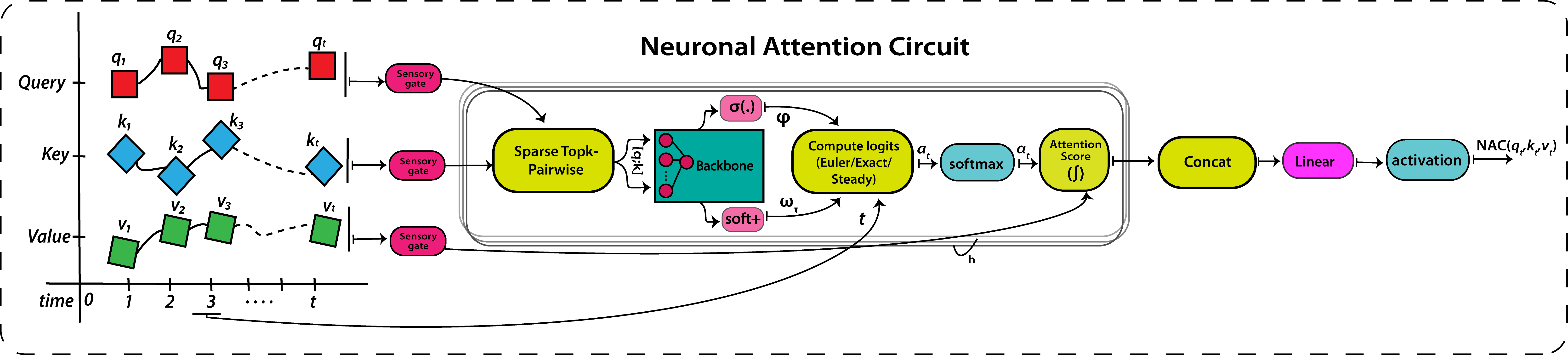}
\caption{Illustration of the architecture of Neuronal Attention Circuit (NAC) mechanism.}
\label{fig:NAC}
\end{figure*}

\begin{algorithm}[t]
\small
\caption{Repurposed NCPCell}
\label{algo:repurposed_NCP}
\begin{algorithmic}
\REQUIRE Wiring $\mathcal{W}$ with $(A_{\text{in}}, A_{\text{rec}})$,
groups $(N_s, N_i, N_c, N_m)$, activation $\alpha$,
input group $G_{\text{input}}$, output group $G_{\text{output}}$,
disabled groups $\mathcal{D}$
\ENSURE Output $y_t \in \mathbb{R}^{B \times d_{\text{out}}}$,
state $\mathbf{x}_t \in \mathbb{R}^{B \times d_h}$
\STATE Binary mask: $M_{\text{rec}} \gets |A_{\text{rec}}|$,
\quad $M_{\text{in}} \gets |A_{\text{in}}|$
\STATE Initialize parameters : $W_{\text{in}}, W_{\text{rec}}, b,$
$w_{\text{in}}, b_{\text{in}}, w_{\text{out}}, b_{\text{out}}$
\STATE Input neurons: $I_{\text{in}} \gets G_{\text{input}}$
\STATE Output neurons:$I_{\text{out}} \gets G_{\text{output}}$
\STATE Define activation mask: $\text{mask}_{\text{act},i} = 0$ if $i\in\mathcal{D}$ else $1$
\STATE Input Projections: $\tilde{u}_t \gets u_t \odot w_{\text{in}} + b_{\text{in}}$
\STATE Recurrent computation: $r_t \gets \mathbf{x}_{t-1}(W_{\text{rec}}\odot M_{\text{rec}})$
\STATE Sparse computation: $s_t \gets \tilde{u}_t (W_{\text{in}}\odot M_{\text{in}})$
\STATE Neuron update: $\mathbf{x}_t \gets
\alpha(r_t + s_t + b)\odot \text{mask}_{\text{act}}$
\STATE Output mapping:  $y_t \gets
(\mathbf{x}_t[I_{\text{out}}]\odot w_{\text{out}}) + b_{\text{out}}$
\STATE \RETURN $(y_t, \mathbf{x}_t)$
\end{algorithmic}
\end{algorithm}

\begin{algorithm}[t]
\small
\caption{Sparse Top-\emph{K} Pairwise Concatenation}
\label{algo:sparse_topk}
\begin{algorithmic}
\REQUIRE Queries $Q \in \mathbb{R}^{B \times H \times T_q \times D}$, Keys $K \in \mathbb{R}^{B \times H \times T_k \times D}$, Top-\emph{K} value $K$
\ENSURE Concatenated tensor $U \in \mathbb{R}^{B \times H \times T_q \times K_{\text{eff}} \times 2D}$
\STATE Block Size: $B_s \gets \lfloor \sqrt{T_k} \rfloor$
\STATE Partition $K$ into $N_b$ blocks: $K_{\text{blocks}} \in \mathbb{R}^{B \times H \times N_b \times B_s \times D}$
\STATE Block Centroids: $\bar{K} \gets \text{mean}(K_{\text{blocks}})$
\STATE Coarse Scores: $S_{\text{coarse}} \gets Q \cdot \bar{K}^\top$
\STATE Select Top-$M$ Blocks: $I_{\text{blocks}} \gets \text{top\_k}(S_{\text{coarse}}, \lceil K/B_s \rceil)$
\STATE Gather Candidates: $K_{\text{cand}} \gets \text{gather}(K_{\text{blocks}}, I_{\text{blocks}})$
\STATE Fine Scores: $S_{\text{fine}} \gets \text{reduce\_sum}(Q \odot K_{\text{cand}})$
\STATE Effective Top-\emph{K}: $K_{\text{eff}} \gets \min(K, \text{size}(K_{\text{cand}}))$
\STATE Indices: $I_{\text{topk}} \gets \text{top\_k}(S_{\text{fine}}, K_{\text{eff}})$
\STATE Selected Keys: $K_{\text{selected}} \gets \text{gather}(K_{\text{cand}}, I_{\text{topk}})$
\STATE Concatenate: $U_{\text{topk}} \gets [\,Q;\; K_{\text{selected}}\,] \in \mathbb{R}^{B \times H \times T_q \times K_{\text{eff}} \times 2D}$
\STATE \RETURN $U_{\text{topk}}$
\end{algorithmic}
\end{algorithm}

\begin{figure}[ht!]
\centering
\includegraphics[width=0.38\textwidth]{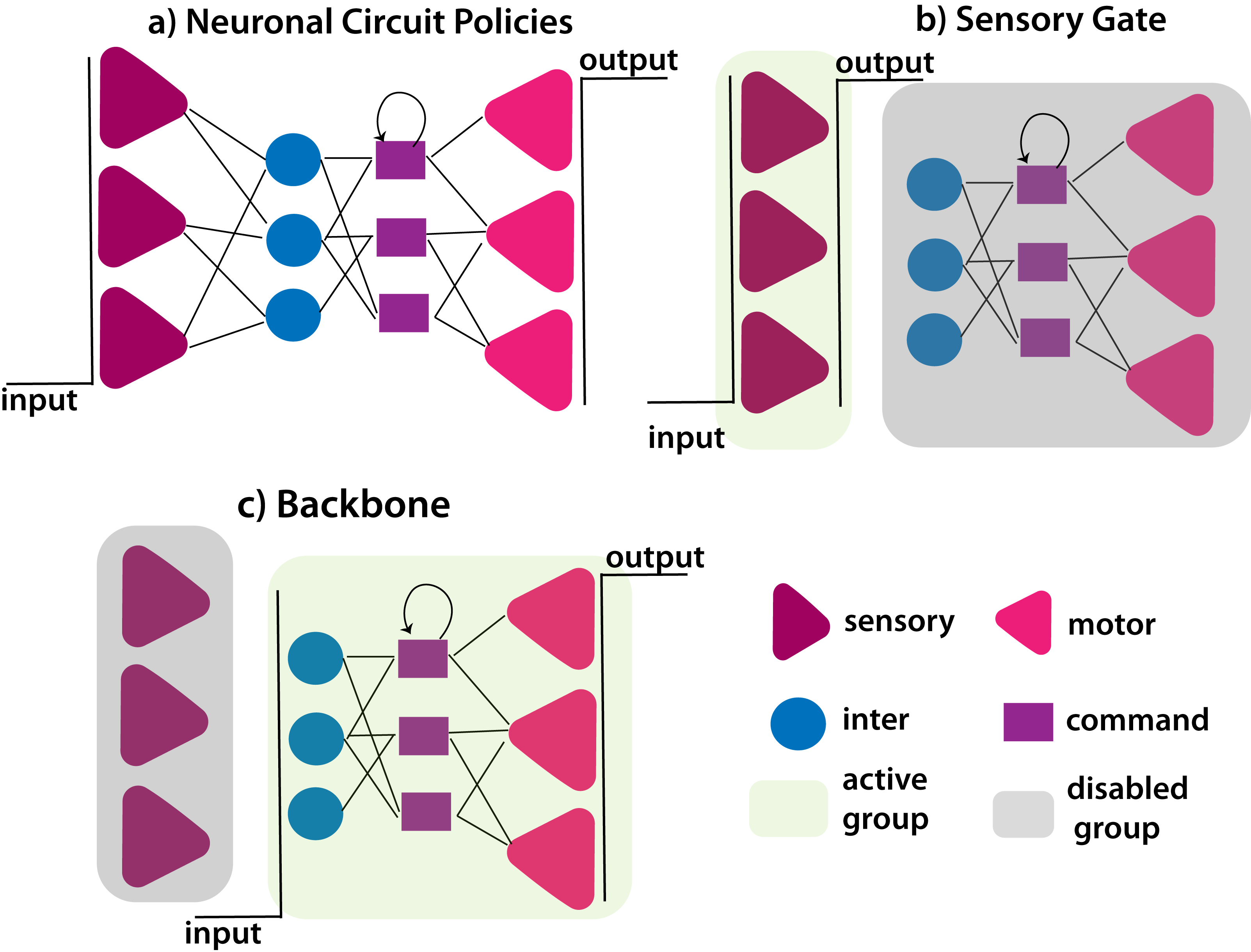}
\caption{Illustration of \textbf{(A)} NCPs with predetermined wiring; \textbf{(B)} Sensory gate, where sensory neurons are active, and the remaining neurons are disabled for the $q$, $k$, and $v$ projections; \textbf{(C)} Backbone, showing inter-motor projections with sensory neurons disabled in extended heads for computing $\phi$ and $\omega_\tau$.}
\label{fig:R_NCPs}
\end{figure}

\subsection{Designing NAC as Neural Network Layer}\label{sec:neural_net_form}
We now outline the design of NAC as neural network layer guided by the preceding analysis. The process involves five steps: (i) repurposing NCPs; (ii) input curation; (iii) construction of the time vector (\(t\)); (iv) computing attention logits and weights; and (v) generating the attention output. Figure~\ref{fig:NAC} provides a graphical overview of NAC.\\
\textbf{Repurposing NCPs:} We repurpose the NCPs framework by converting its fixed, biologically derived wiring (see Figure~\ref{fig:R_NCPs}(a)) into a flexible recurrent architecture that allows configurable input--output mappings. Instead of enforcing a static connectome, our approach exposes adjacency matrices as modifiable structures defining sparse input and recurrent connections. This enables selective information routing across neuron groups while retaining the original circuit topology. Decoupling wiring specifications from model instantiation allows dynamic connectivity adjustments to accommodate different input modalities without full retraining. Algorithm~\ref{algo:repurposed_NCP} summarizes the steps for repurposing the NCPs wiring mechanism. Key features include groupwise masking for neuron isolation, adaptive remapping of inputs and outputs for task-specific adaptation, and tunable sparsity \(s\) to balance expressiveness and efficiency. \\
In our implementation, the sensory neuron gate ($\mathcal{NN}_{\text{sensory}}$) projects the $q$, $k$, and $v$ representations (see Figure~\ref{fig:R_NCPs}(b)). This enables sensory neurons to maintain structured, context-aware representations rather than collapsing inputs into fully connected layers. As a result, the network preserves locality and modularity, which improves information routing. Intuitively, it acts like biological receptor neurons, converting external stimulus inputs into organized signals that feed into downstream circuits.
\begin{equation}
\begin{aligned}
\mathcal{NN}_{\text{sensory}}
&= \text{NCPCell}(G_{\text{input}} = [N_s],\, \\
&\qquad G_{\text{output}} = [N_s], \mathcal{D} = [N_i, N_c, N_m],\, s)
\end{aligned}
\end{equation}
The inter-to-motor pathways form a backbone network ($\mathcal{NN}_{\text{backbone}}$) with branches that compute $\phi$ and $\omega_{\tau}$ (see Figure~\ref{fig:R_NCPs}(c)). Intuitively, this integrates neuronal activity over time and produces outputs that regulate gating signals. Instead of learning $\phi$ and  $\omega_{\tau}$ independently, this backbone allows the model to learn shared representations, enabling multiple benefits: (i) accelerates convergence during training; (ii) separate head layers enable the system to capture temporal and structural dependencies independently.
\begin{equation}
\begin{aligned}
\mathcal{NN}_{\text{backbone}}
&= \text{NCPCell}(G_{\text{input}} = [N_i],\ \\
& \qquad G_{\text{output}} = [N_m],\mathcal{D} = [N_s],\, s)
\end{aligned}
\end{equation}
The output heads are defined as:
\begin{align}
\phi &= \sigma(\mathcal{NN}_{\text{backbone}}(\mathbf{u})) \label{eq:phi_calc} \\
\omega_{\tau} &= \mathrm{softplus}(\mathcal{NN}_{\text{backbone}}(\mathbf{u})) + \varepsilon, \quad \varepsilon > 0 \label{eq:tau_calc}
\end{align}
Here, \(\phi\) serves as a \textit{content–target} gate head, where the sigmoid function \(\sigma(\cdot)\) determines the target signal strength. In contrast, \(\omega_\tau\) is a strictly positive \textit{time–constant} gate head that controls the rate of convergence and the steady-state amplitude. Conceptually, this parallels recurrent gating: \(\phi\) regulates \emph{what} content to emphasize, while \(\omega_\tau\) governs \emph{how quickly} and \emph{to what extent} it is expressed.\\
\textbf{Input Curation:} Initially, full pairwise concatenation was used to form a joint tensor \( U \in \mathbb{R}^{B \times H \times T_q \times T_k \times 2D} \) from $Q \in \mathbb{R}^{B \times H \times T_q \times D}$ and $K \in \mathbb{R}^{B \times H \times T_k \times D}$. However, this approach becomes computationally intractable as the sequence length increases, leading to \( \mathcal{O}(n^2) \) memory growth, severe memory fragmentation, and frequent hardware memory exhaustion. A naive sparse Top-\emph{K} pruning strategy was then implemented to reduce the load on subsequent layers. While this reduced downstream computation, it failed to address the fundamental quadratic bottleneck in the initial scoring and storage stages, rendering the optimization ineffective for long sequences. To mitigate this issue, we then implemented a subquadratic Top-\emph{K} selection strategy based on square-root partitioning ~\cite{child2019sparse}. By dynamically segmenting the key sequence into blocks and scoring queries against block centroids, the method identifies high-salience candidate regions without ever materializing the full \( T_q \times T_k \) matrix. This hierarchical approach reduces the scoring complexity to \( \mathcal{O}(n\sqrt {n}) \), producing a sparse tensor \( U_{\text{topk}} \) that enables linear memory scaling in subsequent backbone layers. The mechanism is self-adaptive and eliminates the need for manual block-size tuning (Algorithm~\ref{algo:sparse_topk}).\\
\textbf{Construction of Time Vector ($t$):} NAC builds on continuous-depth models as in~\cite{hasani2022closed}, which adapt their temporal dynamics to the task. It constructs an internally normalized per-head time vector $t = \sigma(-t_a t_{\mathrm{sample}} + t_b)$, where $t_a$ and $t_b$ are learnable affine parameters and $t_{\mathrm{sample}}$ is derived from the input sample. When the task contains meaningful temporal structure, $t_{\mathrm{sample}}$ introduces input-dependent modulation of the dynamics. When temporal information is not meaningful, $t_{\mathrm{sample}}$ is set to 1, and $t$ reduces to a learned vector through $\sigma(-t_a + t_b)$. In both cases, the $\sigma$ ensures $t \in [0,1]$, providing a smooth, bounded representation of $t$ for modulating the network’s dynamics.\\
\textbf{Attention logits and weights:} Starting from Eqn.~\ref{eq:lq_exact}, consider the trajectory of a query--key pair with initial condition $a_0 = 0$:\\
\begin{equation}
a_t = \frac{\phi}{\omega_\tau} \left(1 - e^{-\omega_\tau t}\right),
\end{equation}
For finite $t_{\text{sample}}$, the exponential factor $(1 - e^{-\omega_\tau t})$ regulates the buildup of attention, giving $\omega_\tau$ a temporal gating role. Normalizing across all keys via \textit{softmax} yields attention weights $\alpha_t = \text{softmax}(a_t)$, defining a valid probability distribution where $\phi$ amplifies or suppresses content alignments, and $\omega_\tau$ shapes both the speed and saturation of these preferences.\\
As $t_{\text{sample}} \to \infty$, the trajectory converges to the steady state
\begin{equation}
a^*_t = \frac{\phi}{\omega_\tau},
\label{eqn:steady}
\end{equation}
which we introduce as a computation mode. Conceptually, it is analogous to SDPA: it aggregates information across inputs in a similar form, even though $\phi$ cannot reproduce exact dot products. \\
\textbf{Attention output:} Finally, the attention output is computed by integrating the \(\alpha_t\) with the \(v_t\) matrix over internally constructed time-window \(t\):
\begin{equation}
\text{NAC}(q, k, v) = \int_{T}\alpha_t v_t dt
\end{equation}
Intuitively, this integral acts as a continuous aggregator of the \(v_t\), where each contribution is weighted by its instantaneous relevance. In practice, we approximate it via a time-modulated sum: \(\sum_i \alpha^{(i)}_t v^{(i)}_t \, w^{(i)}\) where $w^{(i)} \in [0,1]$ is an adaptive input-dependent quadrature weight vector derived from the normalized time vector $t$. Unlike a fixed step size $\Delta t$, $w^{(i)}$ is optimized end-to-end and allows the model to modulate temporal contributions independently of semantic alignment $\alpha_t^{(i)}$, while remaining bounded and differentiable.

\subsubsection{Extension to Multihead} To scale this mechanism to multihead setting, we project the input sequence into $H$ independent subspaces (heads) of dimension $d_\text{model}/H$, yielding query, key, and value tensors $(q^{(h)}, k^{(h)}, v^{(h)})$ for $h \in \{1, \dots, H\}$. For each head, separate concatenated pairs and logits are computed according to Eqns.~\ref{eq:euler},\ref{eq:lq_exact} or \ref{eqn:steady}, followed by \textit{Softmax} normalization to calculate attention weights. The resulting attention weights $\alpha^{(h)}_t$ are then integrated with the value vector $v^{(h)}$,  to produce head-specific attention outputs. Finally, these outputs are concatenated and linearly projected back into the model dimension. This formulation ensures that each head learns distinct dynamic compatibilities governed by its own parameterization of $\phi$ and $\omega_\tau$, aggregation across heads preserves the expressive capacity of the standard multihead attention mechanism.

\subsection{NAC as Universal Approximator}
We now establish the universal approximation capability of NAC by extending the classical Universal Approximation Theorem (UAT) \cite{nishijima2021universal} to the proposed mechanism. For brevity, we consider a network with a single NAC layer processing fixed-dimensional inputs, although the argument generalizes to sequences.
\begin{theorem}[Universal Approximation by NAC]\label{theorem:uat}
Let $K \subset \mathbb{R}^n$ be a compact set and $f : K \to \mathbb{R}^m$ be a continuous function. For any $\epsilon > 0$, there exists a neural network consisting of a single NAC layer, with a sufficiently large model dimension $d_{\text{model}}$, number of heads $H$, sparsity $s$, and nonlinear activations, such that the network’s output $g : \mathbb{R}^n \to \mathbb{R}^m$ satisfies
\begin{equation}
\sup_{x\in K} \|f(x) - g(x)\| < \epsilon.
\end{equation}
The proof is provided in Appendix~\ref{appedix:theorem_uat}.
\end{theorem}

\section{Evaluation}\label{sec:eval}
We evaluate NAC against a range of baselines, including CT-RNNs, (DT \& CT) Attentions, and multiple ablation configurations. Experiments are conducted across diverse domains, including irregular time-series, long-range dependencies, lane keeping of autonomous vehicles, and Industry 4.0 prognostics. All the results are obtained using 5-fold cross-validation, where the models are trained using BPTT (see Appendix~\ref{appendix:training}) on each fold and evaluated across all the folds. We report the mean (\(\mu\)) and standard deviation (\(\sigma\)) to capture variability and quantify uncertainty in the predictions. To ensure best possible hyperparameters we utilized Bayesian optimization~\cite{snoek2012practical} to tune NAC (see Appendix~\ref{appendix:BO}). Table~\ref{tab:results} provides the results for all the experiments, and the details of the data, preprocessing, and neural network architectures for all the experiments are provided in Appendix \ref{appendiX:experiments}.

\subsection{Baselines}
The brief descriptions of baselines are divided into three subcategories.\\
\textbf{CT-RNNs:} CT-RNN \cite{rubanova2019latent} models temporal dynamics using differential equations, enabling hidden states to evolve continuously over time in response to inputs, which is particularly useful for irregularly sampled time-series data. PhasedLSTM \cite{neil2016phased} introduces a time gate that updates hidden states according to a rhythmic schedule, enabling efficient modeling of asynchronous or irregularly sampled time-series. GRU-ODE \cite{de2019gru} extends the GRU to continuous time, evolving hidden states via ODEs to handle sequences with nonuniform time intervals. mmRNN \cite{lechner2022mixed} combines short-term and long-term memory units to capture both fast-changing and slowly evolving patterns in sequential data. LTC \cite{hasani2021liquid} uses neurons with learnable, input-dependent time constants to adapt the speed of dynamics and capture complex temporal patterns in continuous-time data: LTC-FC used FullyConnected layer and LTC-NCP uses NCPs framework. CfC \cite{hasani2022closed} approximate LTC dynamics analytically, providing efficient continuous-time modeling without relying on numerical ODE solvers: CfC-FC used FullyConnected layer and CfC-NCP uses NCPs framework. DeepState~\cite{rangapuram2018deep} combines RNNs with linear state-space models (SSMs), allowing the RNN to produce time-varying SSM parameters to model CT dynamics. S4 \cite{gu2021efficiently} parametrizes SSMs with a structured low-rank correction to the state matrix, enabling efficient convolution-based computation and effective modeling of long-range dependencies.\\
\textbf{DT-Attentions:} SDPA \cite{vaswani2017attention} computes attention weights by measuring the similarity between queries and keys, scaling the results, and applying Softmax to weigh time-step contributions. Linear Attention~\cite{zhang2021sparse} replaces the softmax activation with a \textit{ReLU} function to induce sparsity and enable heads to selectively switch off for certain queries. Performer~\cite{choromanski2020rethinking} approximates Softmax attention with linear-time FAVOR+, enabling efficient and scalable Attention without losing theoretical guarantees.\\
\textbf{CT-Attentions:} mTAN \cite{shukla2021multi} learns continuous-time embeddings and uses time-based attention to interpolate irregular observations into a fixed-length representation for downstream encoder-decoder modeling. CTA \cite{chien2021continuous} generalizes discrete-time attention to continuous-time by representing hidden states, context vectors, and attention scores as functions whose dynamics are modeled via neural networks and integrated using ODE solvers for irregular sequences. ODEFormer \cite{d2023odeformer} trains a sequence-to-sequence transformer on synthetic trajectories to directly output a symbolic ODE system from noisy, irregular time-series data. ContiFormer \cite{chen2023ContiFormer} builds a CT-Transformer by pairing ODE-defined latent trajectories with a time-aware attention mechanism to model dynamic relationships in irregular time-series data. OT-Transformer \cite{kan2025ot} modifies the transformers by interpreting the composition of blocks as an ODE and regularizes training with an optimal transport cost that penalizes the squared norm of the dynamics to improve stability. PDE-Attention \cite{zhang2025continuous} proposes to evolve the attention matrix over a pseudo-time dimension using partial differential equations, smoothing local noise, and enabling polynomial decay of long-range dependencies to improve performance.

\subsection{Ablations Details}\label{appendix:ablation}
The brief descriptions of variants and ablations are divided into five subcategories:\\
\textbf{Top-\emph{K} Ablations:} \textit{NAC-2k} uses Top-\emph{K}=2 to compute the logits and \textit{NAC-32k} uses Top-\emph{K}=32. \textit{NAC-PW} uses fully pairwise (no sparsity) concatenation. All variants use the exact computation mode with 50\% sparsity.\\
\textbf{NAC-Sparsity Ablations:} \textit{NAC-02s} uses 20\% sparsity to compute the logits, and \textit{NAC-09s} uses 90\%. All variants use the exact computation mode with Top-\emph{K}=8. \\
\textbf{NAC with isolated NCPs Ablations:} To isolate the effect of the NCP gating mechanism, we ablate it using MLP layers with matched sparsity. \textit{NAC-FC} replaces the gating module with a standard fully connected layer.  \textit{NAC-05s} applies 50\% sparsity to fully connected layer. \textit{NAC-RW} replaces the gated module with Random Wiring with 50\% sparsity.  All variants use the exact computation mode with Top-\(K = 8\). \\
\textbf{NAC with unconstrained $\phi$:} Applying a sigmoidal nonlinearity to $\phi$ yields the logit to only positive values. To allow negative logits, we tested two variants: (i) \textit{NAC-$\phi_{\text{linear}}$} with no activation, and (ii) \textit{NAC-$\phi_{\text{tanh}}$} with a tanh activation. \\
\textbf{Modes Ablations:} \textit{NAC-Euler} computes attention logits using the explicit Euler integration method. \textit{NAC-Steady} derives attention logits from the steady-state solution of the exact formulation. \textit{NAC~(main)} computes attention logits using the closed-form exact solution. It also overlaps with other ablations, so we combined it into a single one. All modes use Top-\emph{K}=8 and 50\% sparsity. 

\begin{table*}[t]
\centering
\caption{Model Performance across all categories and datasets}
\label{tab:results}
\resizebox{0.9\textwidth}{!}{%
\begin{tabular}{lcccccccccc}
\toprule
\multirow{2}{*}{\textbf{Model}} 
& \multicolumn{2}{c}{\textbf{Irregular Time-Series}} 
& \multicolumn{2}{c}{\textbf{Multivariate LRD}} 
& \multicolumn{2}{c}{\textbf{Lane-Keeping of AVs}} 
& \multicolumn{3}{c}{\textbf{Industry 4.0 }} \\
\cmidrule(lr){2-3} \cmidrule(lr){4-5} \cmidrule(lr){6-7} \cmidrule(lr){8-10}
& \makecell{\textbf{E-MNIST (↑)}} & \textbf{PAR (↑)} 
& \textbf{ETTm1 (↓)} & \textbf{J.Climate (↓)}
& \textbf{CarRacing (↑)} & \makecell{\textbf{Udacity (↓)}} 
& \textbf{PRONOSTIA (↓)} & \textbf{XJTU-SY (↓)} & \textbf{HUST (↓)} \\
\midrule




CT-RNN & 95.18\textsuperscript{\scriptsize $\pm$0.20} 
& 88.71\textsuperscript{\scriptsize $\pm$0.87}
& 0.0315\textsuperscript{\scriptsize $\pm$0.0010}
& \textbf{0.0153\textsuperscript{\scriptsize $\pm$0.0002}} 
& 80.21\textsuperscript{\scriptsize $\pm$0.27}
& 0.0206\textsuperscript{\scriptsize $\pm$0.0013} 
& 44.32\textsuperscript{\scriptsize $\pm$8.69} 
& 26.01\textsuperscript{\scriptsize $\pm$8.74} 
& 39.99\textsuperscript{\scriptsize $\pm$6.33} \\

GRU-ODE 
& 96.04\textsuperscript{\scriptsize $\pm$0.13} 
& 89.01\textsuperscript{\scriptsize $\pm$1.55} 
& 0.0361\textsuperscript{\scriptsize $\pm$0.0013} 
& 0.0286\textsuperscript{\scriptsize $\pm$0.0017} 
& 80.29\textsuperscript{\scriptsize $\pm$0.72}
& 0.0188\textsuperscript{\scriptsize $\pm$0.0016} 
& 45.11\textsuperscript{\scriptsize $\pm$3.19} 
& 31.20\textsuperscript{\scriptsize $\pm$8.69} 
& 43.91\textsuperscript{\scriptsize $\pm$7.10} \\

PhasedLSTM & 95.79\textsuperscript{\scriptsize $\pm$0.14} 
& 88.93\textsuperscript{\scriptsize $\pm$1.08} 
& 0.0371\textsuperscript{\scriptsize $\pm$0.0011} 
& 0.0700\textsuperscript{\scriptsize $\pm$0.0004} 
& 80.35\textsuperscript{\scriptsize $\pm$0.38} 
& 0.0186\textsuperscript{\scriptsize $\pm$0.0015} 
& 44.15\textsuperscript{\scriptsize $\pm$4.80} 
& 35.49\textsuperscript{\scriptsize $\pm$5.54} 
& 38.66\textsuperscript{\scriptsize $\pm$5.55} \\

mmRNN & 95.74\textsuperscript{\scriptsize $\pm$0.27} 
& 88.48\textsuperscript{\scriptsize $\pm$0.46} 
& 0.0377\textsuperscript{\scriptsize $\pm$0.0024} 
& 0.0523\textsuperscript{\scriptsize $\pm$0.0026} 
& 80.13\textsuperscript{\scriptsize $\pm$0.54} 
& 0.0205\textsuperscript{\scriptsize $\pm$0.0027} 
& 48.50\textsuperscript{\scriptsize $\pm$4.60} 
& 27.84\textsuperscript{\scriptsize $\pm$4.05} 
& 40.11\textsuperscript{\scriptsize $\pm$9.56} \\

LTC-FC & 95.25\textsuperscript{\scriptsize $\pm$0.00} 
& 88.12\textsuperscript{\scriptsize $\pm$0.68} 
& 0.0310\textsuperscript{\scriptsize $\pm$0.0004} 
& 0.0161\textsuperscript{\scriptsize $\pm$0.0001} 
& 76.37\textsuperscript{\scriptsize $\pm$3.01} 
& 0.0245\textsuperscript{\scriptsize $\pm$0.0024} 
& 48.14\textsuperscript{\scriptsize $\pm$5.01} 
& 36.83\textsuperscript{\scriptsize $\pm$8.57} 
& 61.82\textsuperscript{\scriptsize $\pm$15.64} \\

CfC-FC & 94.16\textsuperscript{\scriptsize $\pm$0.49} 
& 88.60\textsuperscript{\scriptsize $\pm$0.34} 
& 0.5884\textsuperscript{\scriptsize $\pm$0.3023}
& 0.0169\textsuperscript{\scriptsize $\pm$0.0001} 
& 80.59\textsuperscript{\scriptsize $\pm$0.33} 
& 0.0198\textsuperscript{\scriptsize $\pm$0.0022} 
& 47.78\textsuperscript{\scriptsize $\pm$3.54} 
& 35.51\textsuperscript{\scriptsize $\pm$3.94} 
& 54.09\textsuperscript{\scriptsize $\pm$10.13} \\

LTC-NCP & 94.77\textsuperscript{\scriptsize $\pm$0.26} 
& 88.79\textsuperscript{\scriptsize $\pm$0.15} 
& 0.0294\textsuperscript{\scriptsize $\pm$0.0007} 
& 0.2508\textsuperscript{\scriptsize $\pm$0.1211}
& 79.19\textsuperscript{\scriptsize $\pm$1.55} 
& 0.0228\textsuperscript{\scriptsize $\pm$0.0022} 
& 52.96\textsuperscript{\scriptsize $\pm$6.04} 
& 55.34\textsuperscript{\scriptsize $\pm$16.03} 
& 78.99\textsuperscript{\scriptsize $\pm$16.96} \\

CfC-NCP & 95.63\textsuperscript{\scriptsize $\pm$1.47} 
& 87.24\textsuperscript{\scriptsize $\pm$0.73} 
& 0.0355\textsuperscript{\scriptsize $\pm$0.0063} 
& 0.8972\textsuperscript{\scriptsize $\pm$0.2534} 
& 80.73\textsuperscript{\scriptsize $\pm$0.22} 
& 0.0212\textsuperscript{\scriptsize $\pm$0.0014} 
& 39.13\textsuperscript{\scriptsize $\pm$4.96} 
& 27.08\textsuperscript{\scriptsize $\pm$1.34} 
& 94.35\textsuperscript{\scriptsize $\pm$15.22} \\

DeepState & 96.01\textsuperscript{\scriptsize $\pm$0.07}
& 86.25\textsuperscript{\scriptsize $\pm$0.62}
& 0.0325\textsuperscript{\scriptsize $\pm$0.0007}
& 0.0160\textsuperscript{\scriptsize $\pm$0.0003} 
& 80.25\textsuperscript{\scriptsize $\pm$0.20}
& 0.0175\textsuperscript{\scriptsize $\pm$0.0007}
& 40.86\textsuperscript{\scriptsize $\pm$4.45}  
& 30.42\textsuperscript{\scriptsize $\pm$5.24}  
& 40.88\textsuperscript{\scriptsize $\pm$2.11}  \\

S4 & 95.21\textsuperscript{\scriptsize $\pm$0.07}
& 89.68\textsuperscript{\scriptsize $\pm$0.84}
& 0.2676\textsuperscript{\scriptsize $\pm$0.0017}
& 0.0205\textsuperscript{\scriptsize $\pm$0.0001} 
& 80.32\textsuperscript{\scriptsize $\pm$0.68}
& 0.0183\textsuperscript{\scriptsize $\pm$0.0010}
& 42.88\textsuperscript{\scriptsize $\pm$8.88}  
& 29.49\textsuperscript{\scriptsize $\pm$3.94}  
& 36.20\textsuperscript{\scriptsize $\pm$9.27} \\

SDPA & 95.94\textsuperscript{\scriptsize $\pm$0.15} 
& 88.36\textsuperscript{\scriptsize $\pm$1.06} 
& 0.0353\textsuperscript{\scriptsize $\pm$0.0008}
& 0.0172\textsuperscript{\scriptsize $\pm$0.0003} 
& 79.99\textsuperscript{\scriptsize $\pm$0.49} 
& 0.0185\textsuperscript{\scriptsize $\pm$0.0017} 
& 45.36\textsuperscript{\scriptsize $\pm$5.16} 
& 37.31\textsuperscript{\scriptsize $\pm$12.20} 
& 41.40\textsuperscript{\scriptsize $\pm$7.72} \\

Linear~Attention & 95.87\textsuperscript{\scriptsize $\pm$0.14}
& 88.75\textsuperscript{\scriptsize $\pm$0.85}
& 0.0407\textsuperscript{\scriptsize $\pm$0.0008} 
& 0.0353\textsuperscript{\scriptsize $\pm$0.0003} 
& 80.16\textsuperscript{\scriptsize $\pm$0.30}
& 0.0196\textsuperscript{\scriptsize $\pm$0.0015}
& 58.35\textsuperscript{\scriptsize $\pm$8.67}
& 87.80\textsuperscript{\scriptsize $\pm$19.52}
& 99.38\textsuperscript{\scriptsize $\pm$37.47} \\

Performer & 96.01\textsuperscript{\scriptsize $\pm$0.18}
& 88.71\textsuperscript{\scriptsize $\pm$1.06}
& 0.0349\textsuperscript{\scriptsize $\pm$0.0007} 
& 0.0253\textsuperscript{\scriptsize $\pm$0.0003} 
& 80.11\textsuperscript{\scriptsize $\pm$0.39}
& 0.0180\textsuperscript{\scriptsize $\pm$0.0007}
& 45.37\textsuperscript{\scriptsize $\pm$4.51} 
& 38.05\textsuperscript{\scriptsize $\pm$5.75} 
& 39.06\textsuperscript{\scriptsize $\pm$7.23}  \\

mTAN & 95.97\textsuperscript{\scriptsize $\pm$0.25} 
& 88.08\textsuperscript{\scriptsize $\pm$0.94} 
& 0.4394\textsuperscript{\scriptsize $\pm$0.0066} 
& 0.0312\textsuperscript{\scriptsize $\pm$0.0003} 
& 80.56\textsuperscript{\scriptsize $\pm$0.22} 
& 0.0178\textsuperscript{\scriptsize $\pm$0.0005} 
& 44.41\textsuperscript{\scriptsize $\pm$7.15} 
& 41.34\textsuperscript{\scriptsize $\pm$3.72} 
& 66.29\textsuperscript{\scriptsize $\pm$4.25} \\

CTA & 95.86\textsuperscript{\scriptsize $\pm$0.14} 
& 88.10\textsuperscript{\scriptsize $\pm$1.10}
& \textbf{0.0266\textsuperscript{\scriptsize $\pm$0.0006} }
& 0.0174\textsuperscript{\scriptsize $\pm$0.0001} 
& 80.54\textsuperscript{\scriptsize $\pm$0.40} 
& 0.0197\textsuperscript{\scriptsize $\pm$0.0016} 
& 39.16\textsuperscript{\scriptsize $\pm$3.54} 
& \textbf{25.86}\textsuperscript{\scriptsize $\pm$1.47} 
& 38.41\textsuperscript{\scriptsize $\pm$4.51} \\

ODEFormer & 95.62\textsuperscript{\scriptsize $\pm$0.20} 
& 88.25\textsuperscript{\scriptsize $\pm$0.66} 
& 0.0278\textsuperscript{\scriptsize $\pm$0.0008} 
& 0.0249\textsuperscript{\scriptsize $\pm$0.0002} 
& 80.54\textsuperscript{\scriptsize $\pm$0.40} 
& 0.0190\textsuperscript{\scriptsize $\pm$0.0012} 
& 42.42\textsuperscript{\scriptsize $\pm$6.98} 
& 35.63\textsuperscript{\scriptsize $\pm$9.24} 
& 40.60\textsuperscript{\scriptsize $\pm$6.83} \\

ContiFormer &96.04\textsuperscript{\scriptsize $\pm$0.23} 
& 81.28\textsuperscript{\scriptsize $\pm$0.85} 
& 0.0279\textsuperscript{\scriptsize $\pm$0.0008} 
& 0.0184\textsuperscript{\scriptsize $\pm$0.0002} 
& 80.47\textsuperscript{\scriptsize $\pm$0.50} 
& \textbf{0.0174\textsuperscript{\scriptsize $\pm$0.0013}} 
& 37.82\textsuperscript{\scriptsize $\pm$7.09} 
& 34.71\textsuperscript{\scriptsize $\pm$4.98} 
& 43.81\textsuperscript{\scriptsize $\pm$10.18} \\

OT-Transformer & \textbf{96.30\textsuperscript{\scriptsize $\pm$0.19}} 
& 89.49\textsuperscript{\scriptsize $\pm$0.52} 
& 0.0347\textsuperscript{\scriptsize $\pm$0.0006} 
& 0.0291\textsuperscript{\scriptsize $\pm$0.0002} 
& 80.41\textsuperscript{\scriptsize $\pm$0.32} 
& 0.0196\textsuperscript{\scriptsize $\pm$0.0013} 
& \textbf{37.77\textsuperscript{\scriptsize $\pm$3.45}} 
& 31.38\textsuperscript{\scriptsize $\pm$8.73} 
& 36.85\textsuperscript{\scriptsize $\pm$4.77} \\

PDE-Attention & 95.95\textsuperscript{\scriptsize $\pm$0.22} 
& 89.47\textsuperscript{\scriptsize $\pm$0.70} 
& 0.0377\textsuperscript{\scriptsize $\pm$0.0008} 
& 0.0336\textsuperscript{\scriptsize $\pm$0.0003} 
& 71.53\textsuperscript{\scriptsize $\pm$10.76} 
& 0.0195\textsuperscript{\scriptsize $\pm$0.0004} 
& 40.61\textsuperscript{\scriptsize $\pm$7.69} 
& 35.32\textsuperscript{\scriptsize $\pm$8.23} 
& 39.08\textsuperscript{\scriptsize $\pm$4.67} \\

\midrule

NAC-2k & 95.73\textsuperscript{\scriptsize $\pm$0.07}
& 89.70\textsuperscript{\scriptsize $\pm$0.98}
& 0.0275\textsuperscript{\scriptsize $\pm$0.0006} 
& 0.0151\textsuperscript{\scriptsize $\pm$0.0001} 
& 80.59\textsuperscript{\scriptsize $\pm$0.46}
& 0.0208\textsuperscript{\scriptsize $\pm$0.0015}
& 43.78\textsuperscript{\scriptsize $\pm$2.71}
& 37.43\textsuperscript{\scriptsize $\pm$9.28}
& 40.51\textsuperscript{\scriptsize $\pm$6.61} \\

NAC-32k & 95.15\textsuperscript{\scriptsize $\pm$0.11}
& 89.44\textsuperscript{\scriptsize $\pm$0.87}
& 0.0272\textsuperscript{\scriptsize $\pm$0.0006} 
& \textbf{0.0148\textsuperscript{\scriptsize $\pm$0.0001} }
& 80.38\textsuperscript{\scriptsize $\pm$0.16}
& \textbf{0.0170\textsuperscript{\scriptsize $\pm$0.0007}}
& 49.53\textsuperscript{\scriptsize $\pm$4.89}
& 32.45\textsuperscript{\scriptsize $\pm$10.84}
& 39.17\textsuperscript{\scriptsize $\pm$12.23} \\

NAC-PW & \textbf{96.64}\textsuperscript{\scriptsize $\pm$0.12}
& \textbf{90.18\textsuperscript{\scriptsize $\pm$1.01}} 
& 0.0374\textsuperscript{\scriptsize $\pm$0.0004} 
& 0.0218\textsuperscript{\scriptsize $\pm$0.0002} 
& \textbf{80.72\textsuperscript{\scriptsize $\pm$0.41}} 
& 0.0177\textsuperscript{\scriptsize $\pm$0.0008} 
& \textbf{37.77\textsuperscript{\scriptsize $\pm$2.56}} 
& 28.01\textsuperscript{\scriptsize $\pm$4.93} 
& \textbf{30.14\textsuperscript{\scriptsize $\pm$6.87}} \\
\noalign{\vskip 1pt}
\hdashline
\noalign{\vskip 1pt}
NAC-09s & 95.86\textsuperscript{\scriptsize $\pm$0.11}
& 89.76\textsuperscript{\scriptsize $\pm$0.65}
& 0.0308\textsuperscript{\scriptsize $\pm$0.0013} 
& 0.0150\textsuperscript{\scriptsize $\pm$0.0002} 
& 80.43\textsuperscript{\scriptsize $\pm$0.17}
& 0.0188\textsuperscript{\scriptsize $\pm$0.0013}
& 47.29\textsuperscript{\scriptsize $\pm$5.52}
& 40.40\textsuperscript{\scriptsize $\pm$8.85}
& 44.39\textsuperscript{\scriptsize $\pm$6.82} \\
NAC-02s & 95.51\textsuperscript{\scriptsize $\pm$0.06} 
& 89.47\textsuperscript{\scriptsize $\pm$1.38} 
& 0.0279\textsuperscript{\scriptsize $\pm$0.0007} 
& 0.0149\textsuperscript{\scriptsize $\pm$0.0001} 
& 80.47\textsuperscript{\scriptsize $\pm$0.27} 
& 0.0188\textsuperscript{\scriptsize $\pm$0.0013} 
& 39.43\textsuperscript{\scriptsize $\pm$5.94} 
& 35.59\textsuperscript{\scriptsize $\pm$3.86} 
& 38.90\textsuperscript{\scriptsize $\pm$6.43} \\
\noalign{\vskip 0.3pt}
\hdashline
\noalign{\vskip 0.3pt}
NAC-$\phi_{\text{linear}}$
& 95.89\textsuperscript{\scriptsize $\pm$0.22}
& 89.73\textsuperscript{\scriptsize $\pm$0.89}
& \textbf{0.0273\textsuperscript{\scriptsize $\pm$0.0006} }
 & 0.0154\textsuperscript{\scriptsize $\pm$0.0001} 
& 80.58\textsuperscript{\scriptsize $\pm$0.24}
& 0.0184\textsuperscript{\scriptsize $\pm$0.0024}
& 42.64\textsuperscript{\scriptsize $\pm$3.68}
& 29.53\textsuperscript{\scriptsize $\pm$4.49}
& \textbf{35.89\textsuperscript{\scriptsize $\pm$7.08}} \\
NAC-$\phi_{\text{tanh}}$ 
& 95.70\textsuperscript{\scriptsize $\pm$0.32}
& 89.67\textsuperscript{\scriptsize $\pm$1.27}
& 0.0282\textsuperscript{\scriptsize $\pm$0.0006} 
& 0.0161\textsuperscript{\scriptsize $\pm$0.0001} 
& 80.60\textsuperscript{\scriptsize $\pm$0.06}
& 0.0175\textsuperscript{\scriptsize $\pm$0.0016}
& 45.99\textsuperscript{\scriptsize $\pm$4.55}
& 33.99\textsuperscript{\scriptsize $\pm$3.96}
& 38.75\textsuperscript{\scriptsize $\pm$9.69} \\
\noalign{\vskip 0.3pt}
\hdashline
\noalign{\vskip 0.3pt}
NAC-FC & 95.31\textsuperscript{\scriptsize $\pm$0.07}
& 88.48\textsuperscript{\scriptsize $\pm$1.24}
& 0.0347\textsuperscript{\scriptsize $\pm$0.0007} 
& 0.0294\textsuperscript{\scriptsize $\pm$0.0002} 
& 80.49\textsuperscript{\scriptsize $\pm$0.46}
& 0.0183\textsuperscript{\scriptsize $\pm$0.0017} 
& 47.89\textsuperscript{\scriptsize $\pm$5.16} 
& 41.94\textsuperscript{\scriptsize $\pm$9.57} 
& 46.25\textsuperscript{\scriptsize $\pm$6.83} \\
NAC-05sFC & 95.13\textsuperscript{\scriptsize $\pm$0.14} 
& 88.23\textsuperscript{\scriptsize $\pm$1.44}
& 0.0353\textsuperscript{\scriptsize $\pm$0.0005} 
& 0.0272\textsuperscript{\scriptsize $\pm$0.0001} 
& 75.99\textsuperscript{\scriptsize $\pm$8.81} 
& 0.0263\textsuperscript{\scriptsize $\pm$0.0039} 
& 42.58\textsuperscript{\scriptsize $\pm$4.53} 
& 32.59\textsuperscript{\scriptsize $\pm$10.80} 
& 38.76\textsuperscript{\scriptsize $\pm$9.11} \\

NAC-RW & 95.99 \textsuperscript{\scriptsize $\pm$0.47}
& 88.90\textsuperscript{\scriptsize $\pm$1.43} 
& 0.0314\textsuperscript{\scriptsize $\pm$0.0006} 
& 0.0153\textsuperscript{\scriptsize $\pm$0.0001} 
& 79.75\textsuperscript{\scriptsize $\pm$0.84} 
& 0.0178\textsuperscript{\scriptsize $\pm$0.0010} 
& 45.88\textsuperscript{\scriptsize $\pm$8.44} 
& \textbf{24.55\textsuperscript{\scriptsize $\pm$8.85} }
& 37.79\textsuperscript{\scriptsize $\pm$9.93} \\

\noalign{\vskip 0.3pt}
\hdashline
\noalign{\vskip 0.3pt}

NAC-Euler & 95.67\textsuperscript{\scriptsize $\pm$0.26}
& 89.70\textsuperscript{\scriptsize $\pm$1.54}
& 0.0352\textsuperscript{\scriptsize $\pm$0.0009} 
& \textbf{0.0148\textsuperscript{\scriptsize $\pm$0.0001} }
& \textbf{80.61}\textsuperscript{\scriptsize $\pm$0.28}
& 0.0181\textsuperscript{\scriptsize $\pm$0.0017}
& 42.08\textsuperscript{\scriptsize $\pm$6.14}
& 28.46\textsuperscript{\scriptsize $\pm$8.18}
& 39.32\textsuperscript{\scriptsize $\pm$9.15} \\

NAC-Steady & 95.75\textsuperscript{\scriptsize $\pm$0.28}
& \textbf{89.79\textsuperscript{\scriptsize $\pm$1.67}}
& 0.0266\textsuperscript{\scriptsize $\pm$0.0007} 
& 0.0150\textsuperscript{\scriptsize $\pm$0.0001} 
& \textbf{80.62\textsuperscript{\scriptsize $\pm$0.26}}
& 0.0181\textsuperscript{\scriptsize $\pm$0.0012}
& 40.95\textsuperscript{\scriptsize $\pm$5.77}
& 26.76\textsuperscript{\scriptsize $\pm$7.36}
& 37.12\textsuperscript{\scriptsize $\pm$12.43} \\
\midrule
NAC~(main) & \textbf{96.35}\textsuperscript{\scriptsize $\pm$0.11}
& \textbf{90.11\textsuperscript{\scriptsize $\pm$0.79}}
& \textbf{0.0258\textsuperscript{\scriptsize $\pm$0.0012} }
& \textbf{0.0149\textsuperscript{\scriptsize $\pm$0.0001} }
& 80.59\textsuperscript{\scriptsize $\pm$1.82}
& \textbf{0.0173\textsuperscript{\scriptsize $\pm$0.0006}}
& \textbf{37.75\textsuperscript{\scriptsize $\pm$4.72}}
& \textbf{22.87\textsuperscript{\scriptsize $\pm$1.75}}
& \textbf{27.82\textsuperscript{\scriptsize $\pm$7.09}} \\
\bottomrule
\end{tabular}}
\begin{minipage}{\textwidth}
\footnotesize
\textbf{Note:} (↑) higher is better; (↓) lower is better. \textit{NAC~(main)} uses Exact mode with Top-\emph{K}=8 and 50\% sparsity. 
\end{minipage}
\end{table*}

\subsection{Experiments}
\subsubsection{Irregular Time-series}
We evaluate NAC on two irregular time-series datasets: (i) Event-based MNIST; and (ii) Person Activity Recognition (PAR).\\
\textbf{Event-based MNIST:} Event-based MNIST is the transformation of the widely recognized MNIST dataset with irregular sampling added originally proposed in \cite{lechner2022mixed}. The transformation was performed in two steps: (i) flattening each 28×28 image into a time-series of length 784 and (ii) encoding the binary time-series into an event-based format by collapsing consecutive identical values (e.g., 1,1,1,1 → (1, t=4)). This representation requires models to handle temporal dependencies effectively. NAC-PW achieves an accuracy of 96.64\%, followed by NAC~(main) at 96.35\%. OT-Transformer ranked third with 96.30\%.\\
\textbf{Person Activity Recognition (PAR):} We employed the Localized Person Activity dataset from UC Irvine \cite{localization_data_for_person_activity_196}. The dataset contains data from five participants, each equipped with inertial measurement sensors sampled every 211 ms. The goal of this experiment is to predict a person’s activity from a set of predefined actions, making it a classification task. All models performed well on this task, with NAC-PW achieving 90.18\% accuracy and taking first place. NAC~(main) ranked second with 90.11\% accuracy, while NAC-Steady ranked third with 89.79\% mean accuracy.

\subsubsection{Long-range Dependencies (LRD)}
\begin{figure}[ht!]
\centering
\includegraphics[width=1.0\columnwidth]{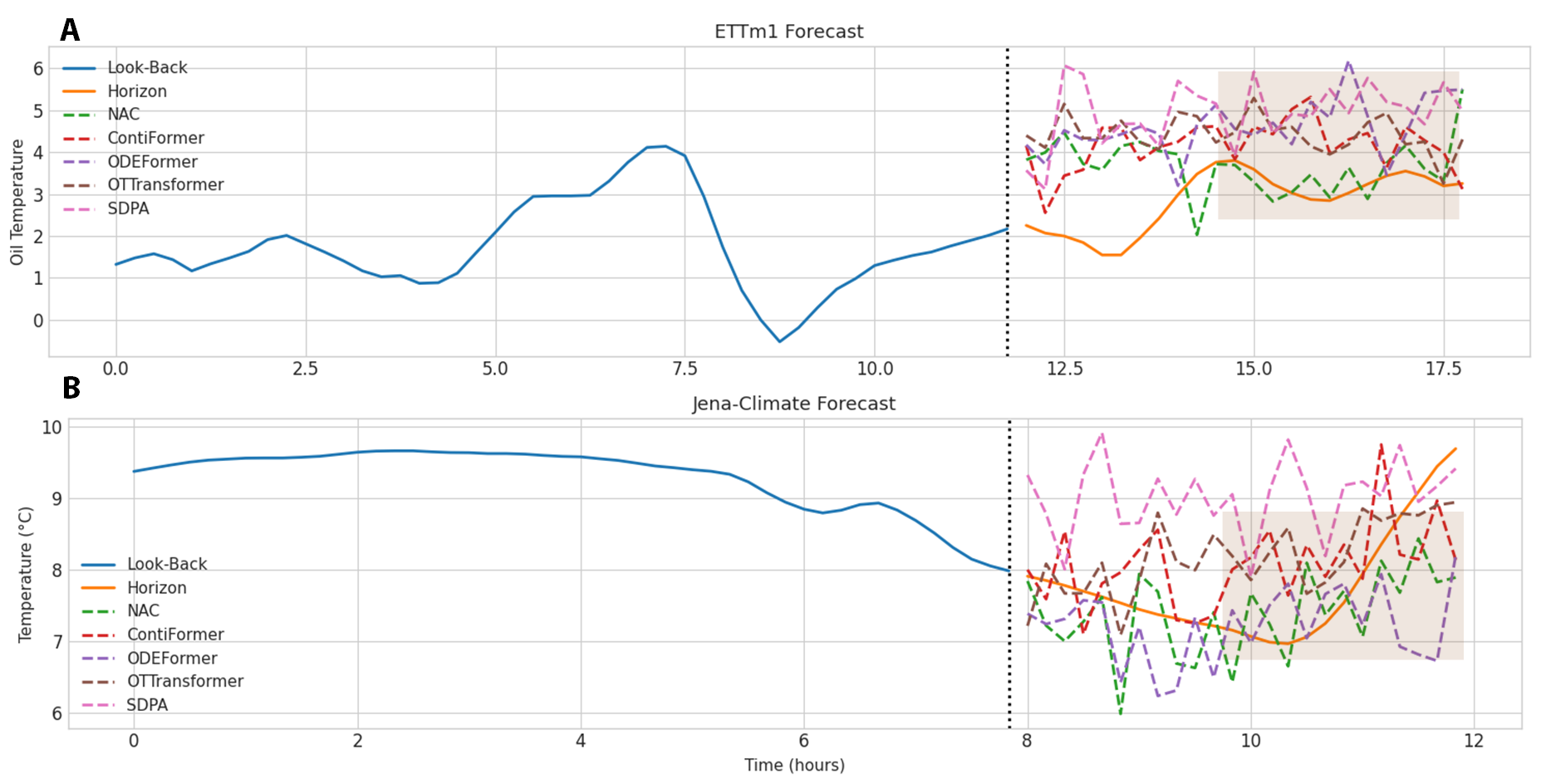}
\caption{Illustration of Forecasting \textbf{(A)} ETTm1 (oil temperature); \textbf{(B)} Jena-Climate (atmosphere temperature).}
\label{fig:forecast}
\end{figure}

To evaluate NAC’s ability to capture LRD, we conduct experiments on two real-world multivariate time series datasets: (i) Electricity Transformer (ETTm1)~\cite{zhou2021informer} and (ii) Jena Climate~\cite{jena_climate_dataset}. Prior to training, both datasets are normalized using \texttt{MinMaxScalar}, and forecasting performance is measured using MSE. We restrict the qualitative illustration of results to Transformer-based models, namely SDPA, Contiformer, ODEFormer, OT-Transformer and NAC~(main) in Figure~\ref{fig:forecast}.\\
\textbf{ETTm1:} ETTm1 consists of electricity transformer measurements recorded at 15-minute intervals, including load and oil temperature variable that capture system dynamics. We train each model on the first 12 hours (48 intervals) of sequence and forecast the subsequent 6 hours (24 intervals). NAC~(main) took first place with the lowest error at 0.0258, followed by CTA at 0.0266, and NAC-$\phi_{\text{linear}}$ at 0.0273. Figure~\ref{fig:forecast}(A) shows that NAC significantly performs better in the highlighted region while competition diverges.\\
\textbf{Jena Climate:} Jena Climate consists of atmospheric variables recorded at 10-minute intervals, including 14 variables such as temperature, pressure, humidity, and wind measurements. We train each model on the first 8 hours (48 intervals) of data and forecast the subsequent 4 hours (24 intervals). NAC-Euler and NAC-32k took first place with the lowest error at 0.0148, followed by NAC~(main) at 0.0149, and CT-RNN at 0.0153. Figure~\ref{fig:forecast}(B) shows the results for Jena-Climate and it shows that NAC performs considerably well against competition.

\subsubsection{Lane Keeping of Autonomous Vehicles}
\begin{figure*}[ht!]
\centering
\includegraphics[width=0.95\textwidth]{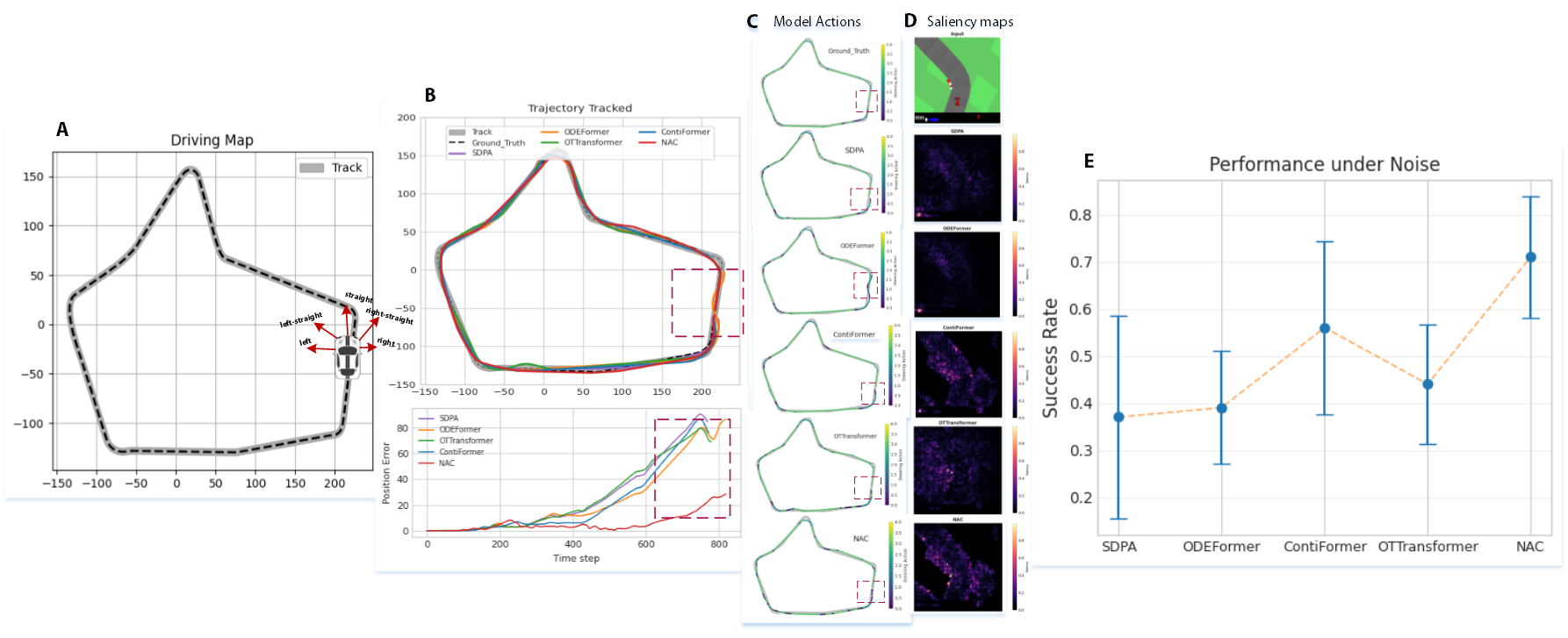}
\caption{Closed-loop test results visualization. \textbf{(A)} driving map; \textbf{(B)} Trajectory tracked and relative position error; \textbf{(C)} Model actions/behavior; \textbf{(D)} Saliency maps; \textbf{(E)} Noise test.}
\label{fig:closed-loop}
\end{figure*}

Lane keeping in autonomous vehicles (AVs) is a fundamental problem in robotics and AI. Early works \cite{tiang2018lane,park2021convolutional} focused primarily on accuracy, often relying on large models. More recent research \cite{lechner2020neural,razzaq2023neural} has shifted toward designing compact architectures suitable for resource-constrained devices. The goal of this experiment is to exploit NAC as a step-wise control module for AVs that creates a temporal structure between the road's horizon and the corresponding steering commands. For evaluation, we used two widely adopted simulation environments: (i) OpenAI-CarRacing \cite{brockman2016openai}; and (ii) the Udacity Self-Driving Car Simulator \cite{udacity}. In OpenAI-CarRacing, the task is to classify steering actions from a predefined action set. In contrast, the Udacity Simulator requires the prediction of a continuous trajectory of steering values. We implemented the AI models proposed by \cite{razzaq2023neural}, replacing the recurrent layer with NAC and its counterparts. All models achieved approximately 80\% accuracy on average in the OpenAI-CarRacing benchmark. Notably, the NAC-PW performed the best, reaching the highest accuracy of 80.72\%, followed by NAC-Steady, ranked second with 80.62\%. NAC-Euler take third position, achieving 80.61\% on average. For the Udacity benchmark, NAC-32k performed the best, achieving the lowest MSE of 0.0170. NAC~(main) followed by 0.0173, and ContiFormer ranked third with 0.0174.\\
Experimental videos are available for the OpenAI-CarRacing \href{https://www.youtube.com/watch?v=kwTNU8aV8-I}{[click here]} and for the Udacity Simulator \href{https://www.youtube.com/watch?v=mMRVsNUQ8i0}{[click here]}.
\textbf{Closed-loop Test:} Beyond prediction accuracy, we performed a closed-loop control experiment on OpenAI-CarRacing to evaluate Transformer models, namely SDPA, ODEFormer, ContiFormer, OT-Transformer, and NAC~(main), to better understand their action dynamics. Figure~\ref{fig:closed-loop}(A) illustrates the driving track. Figure~\ref{fig:closed-loop}(B) shows the trajectories produced by each model, along with the corresponding position error relative to the centerline of the track. In the highlighted region, NAC maintains a substantially lower error and closely follows the centerline, whereas competition deviates from the track, in some cases leaving the track entirely (Figure~\ref{fig:closed-loop}(C)). To further interpret model behavior, we apply Grad CAM~\cite{selvaraju2017grad} to visualize the spatial regions influencing the decision. NAC and ContiFormer consistently attended to the road’s horizon, indicating stable task-relevant focus. In contrast, OT-Transformer exhibits attention drift  outside road boundaries while ODEFormer and SDPA show weaker and less consistent activation, suggesting reduced confidence in salient regions (Figure~\ref{fig:closed-loop}(D). We additionally assess robustness under noise by exploiting the OpenAI-CarRacing built-in stochastic variations, which alter background color and texture, while preserving track geometry. Each model is evaluated over 10 runs of 10 episodes, and we report the mean success rate(Figure~\ref{fig:closed-loop}(E)). NAC achieves the highest performance (70\%), followed by ContiFormer (55\%), with all other methods remaining below 50\%. We attributed NAC’s improved robustness to its dynamical properties, which resemble a low-pass filtering effect and help suppress high-frequency perturbations.

\subsubsection{Industry 4.0}
\begin{figure}[ht!]
\centering
\includegraphics[width=0.85\columnwidth]{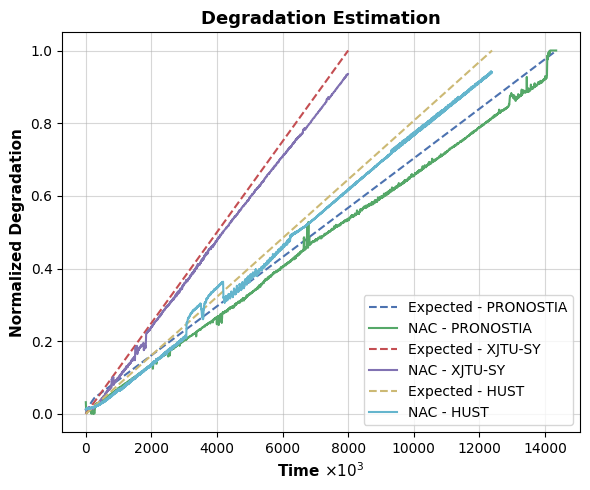}
\caption{Degradation Estimation Results.}
\label{fig:rul}
\end{figure}

Industry 4.0 has revolutionized manufacturing, rendering prognostic health management (PHM) systems indispensable. A key PHM task is estimating the remaining useful life (RUL) of components, particularly rolling element bearings (REB), which account for 40–50\% of machine failures \cite{ding2021remaining,zhuang2021temporal}. The objective is to learn degradation features from one operating condition of a dataset and generalize them to unseen conditions within the same dataset. Furthermore, the model must learn long-term temporal dependencies to accurately capture degradation behavior over time. In addition, it should provide accurate RUL estimation on entirely different datasets while maintaining a compact architecture suitable for resource-constrained devices, thereby supporting localized safety. \\
We utilized three benchmark datasets: (i) PRONOSTIA \cite{nectoux2012pronostia}, (ii) XJTU-SY \cite{wang2018xjtu}, and (iii) HUST \cite{thuan2023hust}. Training is performed on PRONOSTIA, while XJTU-SY and HUST are used to assess zero-shot validation. We used the Score metric \cite{nectoux2012pronostia} to assess the performance. We evaluate generalization using \textit{Bearing 1} from the first operating condition of each dataset. Figure \ref{fig:rul} visualizes the expected degradation alongside the outputs of NAC. For the PRONOSTIA dataset,  NAC~(main) performed the best, with the lowest score of 37.75. NAC-PW and OT-Transformer achieved nearly identical scores, with average values of 37.77 and 37.82, respectively. For the XJTU-SY dataset, NAC~(main) has the lowest score of 22.87. NAC-RW ranked second with 24.55, followed by CTA with 25.86. A similar trend was observed on the HUST dataset, where NAC~(main) achieved first place with a score of 27.82, NAC-PW ranked second with 30.14, and NAC-$\phi_{\text{linear}}$ ranked third with 35.89. These results demonstrated the strong cross-validation adaptability of NAC.

\subsubsection{Interpreting Results}
The optimal NAC configuration depends on the temporal characteristics of each domain. For E-MNIST and PAR, NAC-PW performs best, as each query-key pair carries highly discriminative information and sparsity could remove valuable signals. For ETTm1 and Jena Climate, Exact mode with higher Top-\emph{K} captures slow-forming dependencies across diffuse correlations. In autonomous driving, CarRacing relies mostly on steady-state logits since the CNN front-end captures most spatial features, whereas Udacity driving benefits from NAC-32k to retain peripheral road information from side cameras. Industrial prognostics naturally favor Exact mode, given the monotonic bearing degradation that resembles ODE-like convergence. Across all tasks, NAC-Exact with Top-\emph{K}=8 and 50\% sparsity consistently ranks among the top three configurations.

\begin{figure*}[htbp]
\centering
\includegraphics[width=0.95\textwidth]{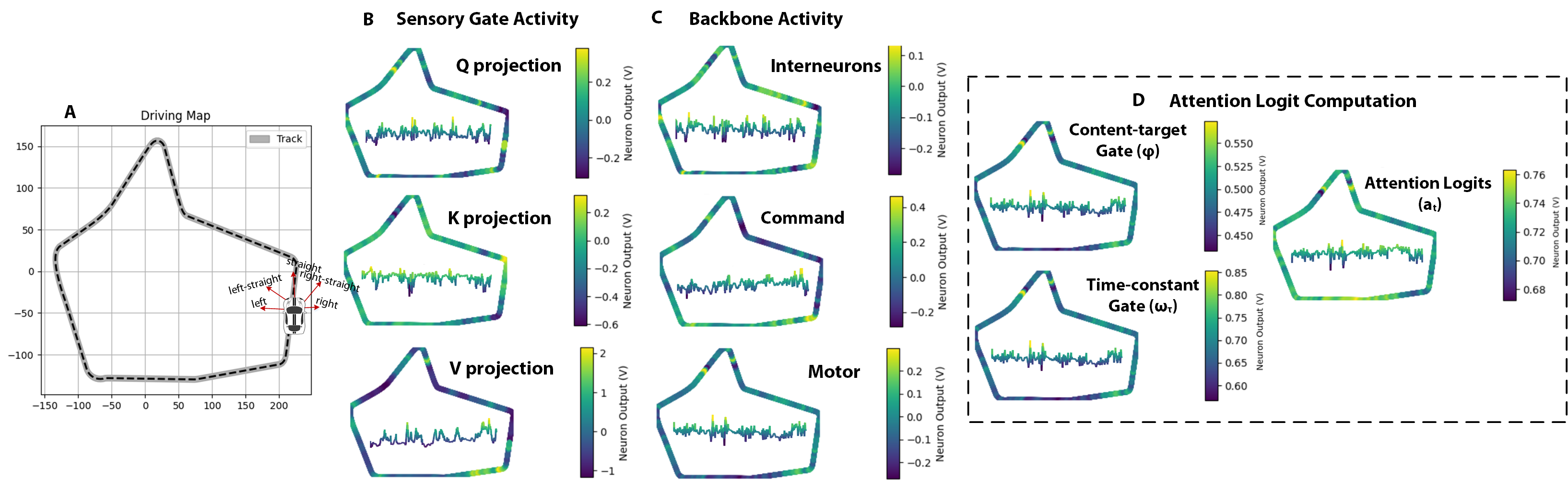}
\caption{Intuitive visualization of NAC cell activity during driving. \textbf{(A)} NAC-controlled car on the driving map. \textbf{(B)} Sensory gate neurons output (centroid) for the $q$, $k$, and $v$ projections. \textbf{(C)} Backbone neuron activity (centroid), including interneurons, command neurons, and motor neurons. \textbf{(D)} \(\omega_\tau\) and \(\phi\) and \(a_t\) computations. Internal plots show actual neuron activities.}
\label{fig:cell_level}
\end{figure*}

\subsection{NAC enhances Interpretability}
Interpretability refers to the process of providing explanations that are understandable to humans, either through visualization or analysis at the neuron cell level. We analyze NAC's neuron outputs at the cell level on OpenAI-CarRacing trained with NAC-Steady, starting from the sensory gates, including the $q$, $k$, and $v$ projections, and extending through the backbone inter-command-motor neurons to the computation of the attention logits. Figure~\ref{fig:cell_level} illustrates the cell-level neuron states of NAC. We observe that \(\omega_\tau\) couples its values between 0.6 and 0.85, depending on the input scene. When the car turns left or left-straight, the \(\omega_\tau\) value remains below 0.65. When it turns right or right-straight, $\omega_\tau$ is maintained above 0.8. In intermediate cases, the model drives straight. This behavior indicates that the evolution of $\omega_\tau$ changes over time, confirming its input dependency. Similarly, the attention logits reflect this behavior, with a tighter range for left and left-straight actions (below 0.70), higher values for right and right-straight actions (above 0.74), and intermediate values corresponding to straight driving. These cell-level neuron states highlight how activations depend on the input, providing a way to understand and evaluate safety-critical decisions.

\begin{table}[ht!]
\centering
\caption{Sequence and time-step prediction complexity. $n$ is the sequence and $k$ is the hidden/model dimension. $S$ is number of ODE solver steps.}
\label{tab:complexity}
\resizebox{0.75\columnwidth}{!}{%
\begin{tabular}{lcc}
\hline
\textbf{Model} & \textbf{Sequence} & \textbf{Time-step} \\
\hline
CT-RNN             & $\mathcal{O}(n k \cdot S)$      & $\mathcal{O}(k \cdot S)$ \\
LNN (ODEsolve) & $\mathcal{O}(nk \cdot S)$ & $\mathcal{O}(k \cdot S)$ \\
SDPA                  & $\mathcal{O}(n^2 k)$   & $\mathcal{O}(nk)$        \\
NAC~(main)     & $\mathcal{O}(n\sqrt{n} k)$   & $\mathcal{O}(nk)$        \\
\bottomrule
\end{tabular}}
\end{table}

\begin{table}[ht!]
\centering
\caption{Run-Time and Memory Benchmark Results}
\label{tab:run_time}
\resizebox{\columnwidth}{!}{%
\begin{tabular}{lccc}
\toprule
\textbf{Model} & \makecell{\textbf{Run-Time}\\(s)} & \makecell{\textbf{Throughput}\\(seq/s)} & \makecell{\textbf{Peak Memory}\\(MB)} \\
\midrule
CT-RNN & 7.1097\textsuperscript{\scriptsize ±0.3048} & 0.141 & 0.67 \\
GRU-ODE & 12.2498\textsuperscript{\scriptsize ±0.0525} & 0.082 & 0.64 \\
PhasedLSTM & 4.9812\textsuperscript{\scriptsize ±0.272} & 0.201 & 0.80 \\
mmRNN & 7.5852\textsuperscript{\scriptsize ±0.2785} & 0.132 & 0.96 \\
LTC-FC & 14.643\textsuperscript{\scriptsize ±0.2445} & 0.068 & 0.99 \\
CfC-FC & 6.0988\textsuperscript{\scriptsize ±0.2135} & 0.164 & 0.76 \\
LTC-NCP & 21.182\textsuperscript{\scriptsize ±0.2383} & 0.05 & 0.56 \\
CfC-NCP & 21.239\textsuperscript{\scriptsize ±0.3327} & 0.05 & 0.57 \\
mTAN & 0.2721\textsuperscript{\scriptsize ±0.0054} & 36.76 & 790.16 \\
ODEFormer & 0.3171\textsuperscript{\scriptsize ±0.0016} & 31.55 & 167.71 \\
CTA & 8.5275\textsuperscript{\scriptsize ±0.2355} & 0.117 & 1.43 \\
ContiFormer & 0.662\textsuperscript{\scriptsize ±0.0075} & 15.15 & 67.71 \\

PDE-Attention & 0.1299\textsuperscript{\scriptsize ±0.0026} & 0.50 & 163.85 \\
OT-Transformer & 0.768\textsuperscript{\scriptsize ±0.0043} & 0.13 & 66.67 \\
\midrule
NAC-2k & 7.3071\textsuperscript{\scriptsize ±0.1547} & 0.137 & 44.75 \\
NAC-32k & 7.2313\textsuperscript{\scriptsize ±0.219} & 0.138 & 549.86 \\
NAC-09s & 7.222\textsuperscript{\scriptsize ±0.176} & 0.139 & 150.85 \\
NAC-02s & 7.252\textsuperscript{\scriptsize ±0.2018} & 0.138 & 151.54 \\
NAC-Euler & 7.3367\textsuperscript{\scriptsize ±0.1719} & 0.136 & 152.22 \\
NAC-Steady & 7.2942\textsuperscript{\scriptsize ±0.1451} & 0.137 & 150.86 \\
\midrule
NAC~(main) & 7.4101\textsuperscript{\scriptsize ±0.1586} & 0.135 & 151.50 \\
\bottomrule
\end{tabular}%
}
\end{table}

\subsection{Analyzing Scalability of NAC}\label{sec:runtime}

\textbf{Efficiency and Complexity: } Table~\ref{tab:complexity} summarizes the computational complexity of different sequence models. For sequence prediction over length $n$ with hidden dimension $k$, CT-RNN and LNN scale $\mathcal{O}(nk\cdot S)$, while SDPA scale quadratically, $\mathcal{O}(n^2 k)$. NAC scale sub quadratically $\mathcal{O}(n\sqrt{n}k)$. For single-time-step prediction, CT-RNNs, LNN require $\mathcal{O}(k\cdot  S)$, whereas SDPA and NAC require $\mathcal{O}(nk)$ when recomputing attention over the full sequence. This places NAC at intermediate position.\\
\textbf{Empirical Analysis: }To analyze the scalability of NAC, we conduct run-time and memory test on fixed-length sequences of 1024 steps, 64-dimensional features, 4 heads, and a batch size of 1. Each model is subjected to 10 forward passes on Google Colab T4-GPU, and we report the mean runtime with standard deviation, throughput, and peak memory usage in Table~\ref{tab:run_time}. NAC occupies an intermediate position in runtime relative to several CT-RNN models, including GRU-ODE, CfCs, and LTCs variants. In terms of memory consumption, NAC uses significantly less memory than mTAN does, with NAC-2k being the least memory-consuming among the NAC ablations. Increasing the NAC sparsity from 20\% to 90\% has a minimal effect on memory, decreasing usage slightly from 151.54~MB to 150.85~MB. In contrast, decreasing the Top-\emph{K} selection from $k=32$ to $k=2$ drastically reduces the memory consumption from 549~MB to 44.75~MB, demonstrating the flexibility of NAC. Compared to LTC-NCP and CfC-NCP architectures, NAC reduces computational cost, achieving approximately a 3× reduction in forward-pass time.

\subsection{Analyzing Impact of Key Hyperparameters}
\begin{figure}[ht!]
\centering
\includegraphics[width=1.0\columnwidth]{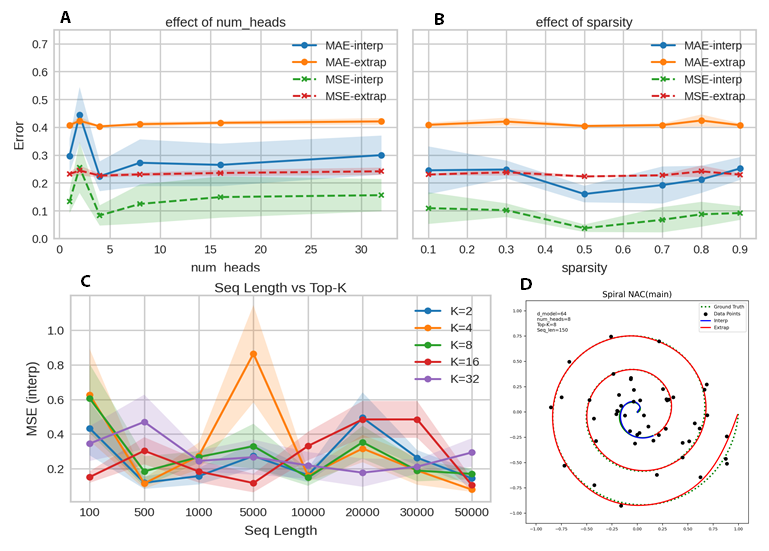}
\caption{Effect of key hyperparameters on interpolation and extrapolation performance (loss curves) for spiral trajectories: \textbf{(A)} heads; \textbf{(B)} sparsity; \textbf{(C)} Seq\_length vs. Top-\emph{K}; \textbf{(D)} Optimized Spiral.}
\label{fig:spiral}
\end{figure}

We now examine how key hyperparameters of NAC including attention heads (1-32), sparsity (0.1-0.9), Top-\emph{K} (2-32) selection relative to sequence lengths (100-50000) affect NAC performance on irregular spiral regression, implemented based on instruction in \cite{chen2023ContiFormer}. All experiments are conducted in Exact mode with a fixed model dimension $d_{\text{model}}=64$. We report both MSE and MAE for interpolation and extrapolation, averaging results over five random initializations and including the corresponding standard deviations. Increasing heads from 1 (no multihead) to 2 slightly reduces performance. Increasing to 4 heads improves accuracy around 37\%, and remains stable (Figure~\ref{fig:spiral}(A)). Changing sparsity has non-monotonic effect, with the lowest error at 50\% (Figure~\ref{fig:spiral}(B)). For Top-\emph{K} selection across sequence length, results are mixed, but 8 $<$ Top-\emph{K} $<$ 16 gives the most consistent performance, though the best value may depends on the task (Figure~\ref{fig:spiral}(C)). Based on these results and Table~\ref{tab:results}, we recommend Top-\emph{K}=8, sparsity=50\%, and Exact mode as the default main NAC configuration. Figure~\ref{fig:spiral}(D) shows interpolation and extrapolation trajectories for this default setup, illustrating accurate and stable reconstructions of the spiral sequences.

\section{Discussions}\label{sec:discuss}
This paper is part of ongoing research on biologically plausible attention mechanisms and represents a pioneering step, with limitations to be addressed in future work.\\
\textbf{Architectural improvement:} Currently, NAC uses predetermined wiring (AutoNCP) requiring three inputs: the number of units (interneuron + motor), motor neurons, and sparsity. To integrate with the attention mechanism while preserving wiring, sensory units for $\mathcal{NN_{\text{sensory}}}$ are set as $\text{units}_\text{sensory} = \left\lceil \frac{d_\text{model} - 0.5}{0.6} \right\rceil$ and backbone units as $\text{units}_\textit{backbone} = d_\text{model} + \left\lfloor \frac{d_\text{model}}{0.6} \right\rfloor$, where $\lceil \cdot \rceil$ and $\lfloor \cdot \rfloor$ denote the ceiling and floor functions, respectively. This results in a larger overall architectural size. Future work will support user-defined NCPs configurations to enable more efficient architectures.\\
\textbf{Neuronal Stochastic Attention Circuit (NSAC):} By reformulating the attention mechanism as a continuous-time evolution process, NAC enables the direct application of Neural ODE frameworks to attention. Extending this formulation by replacing the deterministic ODE with a stochastic differential equation (SDE) is a natural progression, equipping the model with principled uncertainty estimates intrinsic to the attention dynamics.

\section{Conclusion}\label{sec:conclude}
In this paper, we introduce the Neuronal Attention Circuit (NAC), a biologically inspired attention mechanism that reformulates attention logits as the solution to a first-order ODE modulated by nonlinear, interlinked gates derived from repurposing \textit{C.elegans} nematode NCPs. NAC bridges discrete attention with CT dynamics, enabling adaptive temporal processing without the limitations inherent to traditional scaled dot-product attention. Based on the solution to ODE, we introduce three computational modes: (i) Euler, based on \textit{explicit Euler} integration; (ii) Exact, providing closed-form solutions; and (iii) Steady, approximating equilibrium states. In addition, an adaptable sparse Top-\emph{K} pairwise concatenation mechanism is implemented to mitigate the quadratic memory footprint. Theoretically, we establish NAC’s logit-state stability and exponential and frozen approximation error bounds, thereby providing rigorous guarantees of convergence and expressiveness. Empirically, we demonstrated that NAC achieves state-of-the-art performance across diverse tasks, including irregularly time-series, long-range forecasting, autonomous vehicle lane-keeping, and industrial prognostics, while being interpretable at neuron cell-level.


\section*{Reproducibility Statement}
The code is available at:\\
\textbf{Tensorflow:} \url{https://github.com/itxwaleedrazzaq/keras-nac}\\
\textbf{PyTorch:} \url{https://github.com/itxwaleedrazzaq/torch-nac}.




\bibliographystyle{icml2026}
\bibliography{references}

\section*{Appendix}
\appendix

\section{Preliminaries}

\subsection{Neuronal Circuit Policies (NCPs)}\label{appendix:NCPs}
Neuronal Circuit Policies (NCPs) provide a biologically inspired approach for creating interpretable neural control agents, derived from the tap-withdrawal (TW) neural circuit of the nematode \textit{C. elegans} \cite{lechner2018neuronal}. In contrast to conventional spiking neural networks, most neurons in this circuit exhibit electronic dynamics, characterized by the passive propagation of electrical charges and resulting in graded membrane potentials. The NCP architecture is organized into a four-layer hierarchy: sensory neurons (\(N_s\)), touch interneurons (\(N_i\)), command interneurons (\(N_c\)), and motor neurons (\(N_m\)). The sensory neurons (\(N_s\)) detect external environmental states and initiate signal processing. For each input state variable \(x\), an \(N_s\) includes positive (\(S_p\)) and negative (\(S_n\)) sub-neurons, with their membrane voltages modulated based on the sign and magnitude of \(x\): \(S_p\) depolarizes for \(x > 0\) (remaining at rest for \(x \leq 0\)), while \(S_n\) depolarizes for \(x < 0\) (remaining at rest for \(x \geq 0\)). This input \(x\) is nonlinearly mapped to membrane potentials in the range \([-70\,\text{mV}, -20\,\text{mV}]\), aligning with biological neuron properties where \(-70\,\text{mV}\) represents the resting potential and \(-20\,\text{mV}\) approximates the peak activation level. Similarly, each motor neuron (\(N_m\)) comprises positive (\(M_p\)) and negative (\(M_n\)) sub-neurons, whose membrane potentials are inversely mapped to produce the output action \(y\) (as \(y = y_p + y_n\)), ensuring outputs remain within task-specific bounds while preserving biological voltage scales. The NCP wiring emulates the sparse and modular nature of biological neural circuits, featuring feedforward connections from \(N_s\) to \(N_i\), recurrent interactions primarily among \(N_c\) (enabling competitive dynamics for action selection), and targeted projections from \(N_c\) to \(N_m\) \cite{lechner2018neuronal}. Figure \ref{fig:R_NCPs}(a) depicts the NCP connectome.

\subsection{SDPA Mechanism}\label{appendix:attention}
SDPA mechanisms have become a cornerstone in modern neural architectures, enabling models to dynamically focus on relevant parts of the input. The concept was first introduced in the context of neural machine translation, where it allows the decoder to weight encoder outputs according to their importance for generating each target token. Given a query vector \(q \in \mathbb{R}^d\), key vectors \(K = [k_1, k_2, \dots, k_n] \in \mathbb{R}^{n \times d}\), and value vectors \(V = [v_1, v_2, \dots, v_n] \in \mathbb{R}^{n \times d}\), the attention mechanism can be expressed in two steps:
\begin{enumerate}
    \item Compute the scaled dot attention logits:
    \begin{equation}
    a_i = \frac{q^T k_i}{\sqrt{d}}
    \label{eq:attention_logit}
    \end{equation}

    \item Normalize the logits to obtain attention weights and compute the output:
    \begin{align}
    \alpha_i = \text{softmax}(a_i) &=  \frac{e^{a_i}}{\sum_{j=1}^n e^{a_j}}\\
    \text{Attention}(q,k,v) &= \sum_{i=1}^n \alpha_i v_i
    \end{align}
\end{enumerate}
Here, \(a_i\) is the raw attention logit between the query and each key, and the scaling factor \(\sqrt{d}\) prevents large dot products from destabilizing the Softmax  \cite{vaswani2017attention}.

\section{Proofs}
In this section, we provide all the proofs.
\subsection{Analyzing Closed-form Solution}\label{appendix:analyzing_exact}
\subsubsection{Deriving Closed-form (Exact) Solution}\label{appendix:derving_exact}
Although $\phi$ and $\omega_\tau$ are nonlinear functions of the input $\mathbf{u}=[\mathbf{q};\mathbf{k}]$, we derive a closed-form solution by treating them as locally constant over the small integration interval for each query--key pair based on frozen-coefficient approximation~\cite{john1952integration}. Rewrite Eqn.~\ref{eq:lq_diff} as
\begin{equation}
\frac{d a_t}{dt} + \omega_\tau a_t = \phi.
\end{equation}
The integrating factor is
\begin{equation}
\mu = e^{\Big(\int \omega_\tau\, dt\Big)} = e^{\omega_\tau t}.
\end{equation}
Multiply both sides by \(\mu(t)\):
\begin{equation}
e^{\omega_\tau t}\frac{d a_t}{dt} + \omega_\tau e^{\omega_\tau t} a_t = \phi e^{\omega_\tau t}.
\end{equation}
Recognizing the left-hand side as the derivative of \(e^{\omega_\tau t} a_t\):
\begin{equation}
\frac{d}{dt}\big(e^{\omega_\tau t} a_t\big) = \phi e^{\omega_\tau t}.
\end{equation}
Integrate from \(0\) to \(T\):
\begin{equation}
e^{\omega_\tau t} a_t - e^{0} a_0 = \phi \int_0^t e^{\omega_\tau s}\, ds.
\end{equation}
Compute the integral:
\begin{equation}
\int_0^t e^{\omega_\tau s}\, ds = \frac{1}{\omega_\tau}\big(e^{\omega_\tau t} - 1\big).
\end{equation}
Substitute back:
\begin{equation}
e^{\omega_\tau t} a_t - a_0 = \phi \cdot \frac{e^{\omega_\tau t} - 1}{\omega_\tau}.
\end{equation}
Rearrange:
\begin{equation}
e^{\omega_\tau t} a_t = a_0 + \frac{\phi}{\omega_\tau}\big(e^{\omega_\tau t} - 1\big).
\end{equation}
Divide both sides by \(e^{\omega_\tau t}\):
\begin{equation}
a_t = a_0 e^{-\omega_\tau t} + \frac{\phi}{\omega_\tau}\big(1 - e^{-\omega_\tau t}\big).
\label{eq:frozen}
\end{equation}
Set \(a^* := \dfrac{\phi}{\omega_\tau}\). Then \(a_t = a^* + (a_0 - a^*) e^{-\omega_\tau t}\), is proved.

\subsubsection{Frozen Coefficient Error Analysis}

In this subsection, we analyze the resulting error due to the frozen approximation with respect to the time-varying behavior of $\phi$ and $\omega_\tau$, both mathematically and empirically.

\begin{theorem}[Frozen-Coefficient Approximation Error]\label{thm:frozen-error}
Consider the attention logit ODE
\begin{equation}
\frac{da(t)}{dt} = -\omega_\tau(t) a(t) + \phi(t),
\label{eq:attention-ode}
\end{equation}
with time-varying coefficients \(\phi(t)\) and \(\omega_\tau(t) > 0\). Let \(a_{\text{varying}}(t)\) denote the solution with varying coefficients (numerically integrated) and \(a_{\text{frozen}}(t)\) denote the exact closed-form solution obtained by freezing the coefficients at their initial values \(\phi_0 = \phi(0)\), \(\omega_{\tau 0} = \omega_\tau(0)\) over an integration window [0, T]. Assume that \(\phi(t)\) and \(\omega_\tau(t)\) are Lipschitz continuous in t with constants \(L_\phi\) and \(L_\omega\) (i.e., \(|\phi(t) - \phi(s)| \le L_\phi |t - s|\), \(|\omega_\tau(t) - \omega_\tau(s)| \le L_\omega |t - s|\)). Define \(C = |a_0| + |\phi_0| / \omega_{\tau 0}\). Then, for any \(t \in [0, T]\), the error \(\varepsilon(t) = a_{\text{varying}}(t) - a_{\text{frozen}}(t)\) satisfies
\begin{equation}
|\varepsilon(t)| \le \frac{L_\phi + C L_\omega}{L_\omega} \left( e^{L_\omega t^2 / 2} - 1 \right),
\label{eq:error-bound-main}
\end{equation}
For small \(L_\omega t^2 \ll 1\), this simplifies to $\mathcal{O}(t^2)$.
\end{theorem}

\begin{proof}
The frozen-coefficient solution satisfies
\begin{equation}
\frac{da_{\text{frozen}}}{dt} = -\omega_{\tau 0} a_{\text{frozen}} + \phi_0, \quad a_{\text{frozen}}(0) = a(0),
\label{eq:frozen-ode}
\end{equation}
Subtracting Eqn.~\ref{eq:frozen-ode} from Eqn.~\ref{eq:attention-ode} yields the error dynamics
\begin{equation}
\frac{d\varepsilon}{dt} = -\omega_{\tau 0} \varepsilon + \delta\phi(t) + \delta\omega_\tau(t) a_{\text{varying}}(t),
\label{eq:error-dynamics}
\end{equation}
where \(\delta\phi(t) = \phi(t) - \phi_0\) and \(\delta\omega_\tau(t) = \omega_{\tau 0} - \omega_\tau(t)\). By Lipschitz, \(|\delta\phi(t)| \le L_\phi t\) and \(|\delta\omega_\tau(t)| \le L_\omega t\). Using the integrating factor \(e^{\omega_{\tau 0} t}\),
\begin{equation}
\frac{d}{dt} \left( e^{\omega_{\tau 0} t} \varepsilon(t) \right) = e^{\omega_{\tau 0} t} \left( \delta\phi(t) + \delta\omega_\tau(t) a_{\text{varying}}(t) \right),
\label{eq:error-integral-factor}
\end{equation}
Integrating from 0 to $T$ and noting that \(\varepsilon(0) = 0\),
\begin{equation}
\varepsilon(t) = \int_0^t e^{-\omega_{\tau 0} (t-s)} \left( \delta\phi(s) + \delta\omega_\tau(s) a_{\text{varying}}(s) \right) ds
\label{eq:error-integral}
\end{equation}
The frozen solution from Eqn.~\ref{eq:frozen} is
\begin{equation}
a_{\text{frozen}}(s) = a_0 e^{-\omega_{\tau 0} s} + \frac{\phi_0}{\omega_{\tau 0}} \left( 1 - e^{-\omega_{\tau 0} s} \right),
\label{eq:frozen-solution}
\end{equation}
so that \(|a_{\text{frozen}}(s)| \le C\). Hence \(|a_{\text{varying}}(s)| \le C + |\varepsilon(s)|\). Taking absolute values in Eqn.~\ref{eq:error-integral} and using Lipschitz bounds,
\begin{equation}
|\varepsilon(t)| \le \int_0^t e^{-\omega_{\tau 0} (t-s)} \left( L_\phi s + L_\omega s (C + |\varepsilon(s)|) \right) ds,
\label{eq:error-inequality}
\end{equation}
Since \(e^{-\omega_{\tau 0} (t-s)} \le 1\),
\begin{equation}
|\varepsilon(t)| \le \frac{1}{2} (L_\phi + C L_\omega) t^2 + L_\omega \int_0^t s |\varepsilon(s)| ds
\end{equation}
Applying Gronwall's inequality to \(|\varepsilon(t)|\) yields
\begin{equation}
|\varepsilon(t)| \le \frac{L_\phi + C L_\omega}{L_\omega} \left( e^{L_\omega t^2 / 2} - 1 \right),
\end{equation}
which is Eqn.~\ref{eq:error-bound-main}. Expanding for $L_\omega t^2/2 \ll 1$ gives the leading term $\frac{1}{2} (L_\phi + C L_\omega) t^2 = \mathcal{O}(t^2)$.
\end{proof}

\begin{figure}[ht!]
\centering
\includegraphics[width=0.45\textwidth]{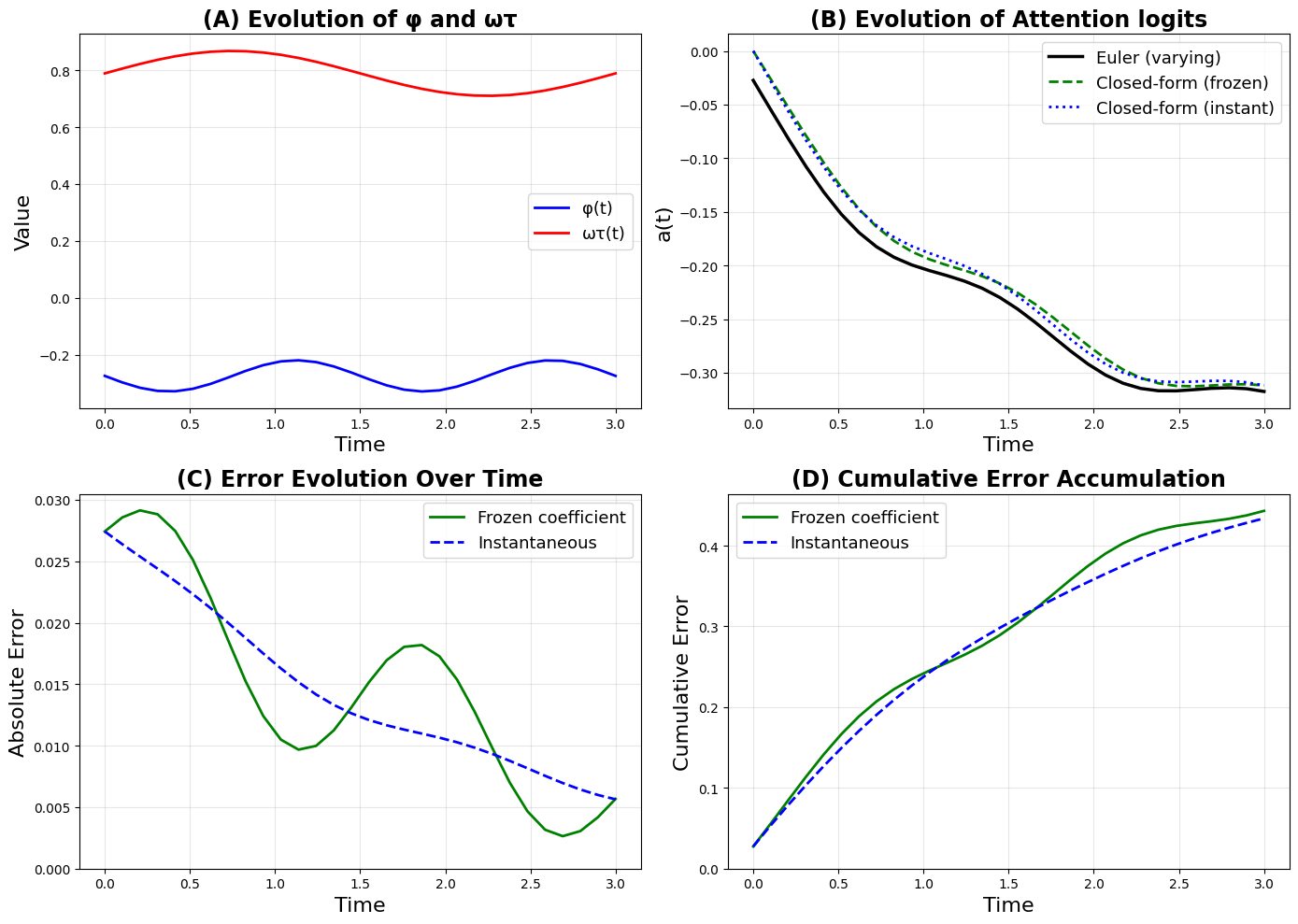}
\caption{Visualization of oscillatory evolution of NAC.}
\label{fig:frozen}
\end{figure}

\textbf{Empirical Analysis:} We now empirically analyze the error introduced by freezing $\phi(t)$ and $\omega_\tau(t)$. We evaluate the closed-form frozen solution against continuously varying coefficients. Empirical analysis across four regimes: (i) slow variation (gradual linear change); (ii) fast variation (exponential growth); (iii) step changes (abrupt discontinuities); and (iv) oscillatory behavior (periodic modulation) shows that the frozen approximation maintains bounded error even under substantial parameter variation. Table~\ref{tab:frozen-empirical} summarizes the empirical results. For typical slow-varying irregular sequences, with $\phi$ varying by 0.5\% and $\omega_\tau$ varying by 0.26\%, the frozen solution exhibits minimal error, with a mean of 1.87\%, and a maximum of 3.05\%. Even under more challenging regimes, the error remains predictable and bounded, reaching 6.41\% for fast variation, 6.53\% for step changes, and 7.81\% for oscillatory behavior. An illustration of oscillatory evolution is shown in Figure~\ref{fig:frozen}.\\
\emph{Interpretation:} Given that the variation rates in real-world irregular sequences rarely exceed 10\%, these results confirm the adequacy of the frozen approximation for practical deployment. This conclusion is further supported by the real-data mode ablation results in Table~\ref{tab:results}, where the \textit{Exact} mode and the numerically integrated Euler variant achieve nearly identical performance on the Jena Climate benchmark (MSE: 0.0149 vs.\ 0.0148) and comparable results across the remaining datasets. The absence of a consistent advantage for Euler integration indicates that the approximation error introduced by coefficient freezing is negligible in practice. We therefore recommend Exact as the default computation mode for NAC while preserving adaptive Euler integration for edge cases with exceptionally rapid dynamics.

\begin{table}[h]
\centering
\caption{Empirical error analysis of frozen-coefficient approximation across variation regimes.}
\label{tab:frozen-empirical}
\begin{tabular}{lcccc}
\hline
\makecell{\textbf{Variation}} & \makecell{\textbf{mean}\\(\%)} & \makecell{\textbf{max}\\(\%)} & \makecell{\boldmath{$\phi$}\\(\%)}  & \makecell{\boldmath{$\omega_\tau$}\\(\%)} \\
\hline
Slow         & 1.87 & 3.05 & 0.50 & 0.26 \\
Fast    & 6.41 & 6.50 & 2.63 & 1.58 \\
Step  & 6.53 & 7.20 & 3.51 & 2.11 \\
Oscillatory & 7.81 & 6.90 & 8.28 & 2.10 \\
\hline
\textbf{Average}       & \textbf{5.65} & \textbf{5.18} & \textbf{3.73} & \textbf{1.51} \\
\hline
\end{tabular}
\end{table}

\subsection{Proof of Theorem \ref{theorem:state_stability}}\label{appendix:theorem_state}
Following the footprints of~\cite{hasani2021liquid}, we divide the proof into two parts: (i) the single-connection case $M=1$, and (ii) the general multi-connection case $M>1$. The main technique is to evaluate the ODE at boundary values of the proposed invariant interval and show that the derivative points inward, ensuring that trajectories cannot escape.

\subsubsection*{Case 1: Single connection ($M=1$).}  
The ODE reduces to
\begin{equation}
\frac{da}{dt} = -\omega_\tau a + \phi = -\omega_\tau(a - A), \quad A = \frac{\phi}{\omega_\tau}.
\end{equation}
Here $A$ is the unique equilibrium. We now check both bounds.

- \emph{Upper bound:} Let $M = \max(0, A)$.  
  At $a=M$,
  \begin{equation}
  \frac{da}{dt}\Big|_{a=M} = -\omega_\tau (M - A).
  \end{equation}
  If $A \geq 0$, then $M=A$ and $\frac{da}{dt}=0$.  
  If $A < 0$, then $M=0$, so $\frac{da}{dt} = -\omega_\tau (0 - A) = \omega_\tau A \leq 0$ since $A<0$.  
  In both cases, $\frac{da}{dt} \leq 0$. Thus, trajectories cannot cross above $M$.

- \emph{Lower bound:} Let $m = \min(0, A)$.  
  At $a=m$,
  \begin{equation}
  \frac{da}{dt}\Big|_{a=m} = -\omega_\tau (m - A).
  \end{equation}
  If $A \leq 0$, then $m=A$ and $\frac{da}{dt}=0$.  
  If $A > 0$, then $m=0$, so $\frac{da}{dt} = -\omega_\tau (0 - A) = \omega_\tau A \geq 0$.  
  In both cases, $\frac{da}{dt} \geq 0$. Thus trajectories cannot cross below $m$. Therefore, the interval $[m, M]$ is forward-invariant.  

To see this explicitly under Euler discretization with step size $\Delta t > 0$,
\begin{equation}
a(t+\Delta t) = a_t + \Delta t \cdot \frac{da}{dt}.
\end{equation}
At $a=M$, $\frac{da}{dt}\leq 0 \implies a(t+\Delta t)\leq M$.  
At $a=m$, $\frac{da}{dt}\geq 0 \implies a(t+\Delta t)\geq m$.  
By induction over steps, $a_t\in[m,M]$ for all $t\in[0,T]$.  

\subsubsection*{Case 2: Multiple connections ($M>1$).}  
The ODE is
\begin{equation}
\frac{da}{dt} = -\Big(\sum_{j=1}^M f_j\Big)a + \sum_{j=1}^M f_j A_j,
\end{equation}
with per-connection equilibria $A_j = \phi_j / f_j$. The effective equilibrium is
\begin{equation}
A = \frac{\sum_{j=1}^M f_j A_j}{\sum_{j=1}^M f_j}.
\end{equation}
Since the weights $\frac{f_j}{\sum f_j}$ are positive and sum to 1, $A$ is a convex combination of $\{A_j\}$. Therefore,
\begin{equation}
A \in [A^{\min}, A^{\max}].
\end{equation}

- \emph{Upper bound:} Let $M = \max(0, A^{\max})$. Then
  \begin{equation}
  \frac{da}{dt}\Big|_{a=M} = \sum_{j=1}^M f_j (A_j - M).
  \end{equation}
  Since $A_j \leq A^{\max} \leq M$, each term $(A_j - M)\leq 0$, and thus $\sum f_j (A_j - M)\leq 0$.  
  Hence $\frac{da}{dt}\leq 0$, proving trajectories cannot exceed $M$.

- \emph{Lower bound:} Let $m = \min(0, A^{\min})$. Then
  \begin{equation}
  \frac{da}{dt}\Big|_{a=m} = \sum_{j=1}^M f_j (A_j - m).
  \end{equation}
  Since $A_j \geq A^{\min} \geq m$, each $(A_j - m)\geq 0$, so $\sum f_j (A_j - m)\geq 0$.  
  Hence $\frac{da}{dt}\geq 0$, proving trajectories cannot fall below $m$. Thus, the interval $[m,M]$ is forward-invariant.  

\begin{remark}
This result guarantees that the continuous-time attention state converges within a well-defined interval dictated by the per-connection equilibria. In particular, for the single-connection case ($M=1$), the state trajectory converges monotonically toward the closed-form equilibrium solution (Eqn.~\ref{eqn:steady}) without overshoot. 
\end{remark}

\subsection{Proof for Theorem \ref{theorem:uat}} \label{appedix:theorem_uat}

The proof proceeds constructively by showing that the NAC layer can emulate a single-hidden-layer feedforward neural network with nonlinear activations, which is a universal approximator under the Universal Approximation Theorem (UAT). We assume self-attention on a single-token input $x \in \mathbb{R}^n$ (setting sequence length $T = 1$) and focus on the steady mode for simplicity. Without loss of generality, set $d_{\text{model}} = n + m$ or adjust as needed for dimensionality. Constructively, set NCP sparsity $s=0$ for full connectivity, ensuring the backbone $\mathcal{NN}_\text{backbone}$ approximates any $\tilde{\phi}: \mathbb{R}^{2d} \to [0,1]$ with error $<\delta$ via stacked layers. For multi-head, scale $H$ proportionally to target complexity, with output projection $W_o$ aggregating as in classical UAT proofs \cite{stinchcomb1989multilayered}.\\
\textbf{Input Projections:} The input $x$ is projected via NCP-based sensory projections to obtain query $q = q_{\text{proj}}(x)$, key $k = k_{\text{proj}}(x)$, and value $v = v_{\text{proj}}(x)$, each in $\mathbb{R}^{d_{\text{model}}}$. For emulation, set $q_{\text{proj}} = k_{\text{proj}} = I_n$ (identity on $\mathbb{R}^n$) and adjust \( W_v \) later, so \( q = k = x \) and \( v \) is configurable. These are affine transformations and do not limit expressivity.\\
\textbf{Head Splitting and Sparse Top-\emph{k} Pairwise Computation:} Split into \( H \) heads, yielding \( q^{(h)}, k^{(h)} \in \mathbb{R}^{d} \) per head \( h \), where \( d = d_{\text{model}} / H \). For \( T=1 \), compute sparse top-\emph{k} pairs, but since \( T=1 \), \( K_{\text{eff}}=1 \), yielding concatenated pair \( u^{(h)} =  [q^{(h)}; k^{(h)}] \in \mathbb{R}^{2d} \). Since \( q^{(h)} = k^{(h)} \), this is \( [x^{(h)}; x^{(h)}] \), but the NCP processes it generally.\\
\textbf{Computation of \( \phi^{(h)} \) and \( \omega_\tau^{(h)} \):} The scalar \( \phi^{(h)} \) is computed via the NCP-based inter-to-motor projection on the pair:
\begin{equation}
\phi^{(h)} = \sigma(\mathcal{NN}_{\text{backbone}}(u^{(h)}))
\end{equation}
where \( \sigma(z) = (1 + e^{-z})^{-1} \) is the sigmoid. This NCP, with sufficiently large units and low sparsity, approximates any continuous scalar function \( \tilde{\phi}: \mathbb{R}^{2d} \to [0,1] \) to arbitrary precision on compact sets (by the UAT for multi-layer networks \cite{stinchcomb1989multilayered}). Similarly, \( \omega_\tau^{(h)} \) is computed via:
\begin{equation}
\omega_{\tau}^{(h)} = \mathrm{softplus}(\mathcal{NN}_{\text{backbone}}(u^{(h)})) + \varepsilon, \quad \varepsilon > 0
\end{equation}
By setting weights to make \( \omega_\tau^{(h)} \equiv 1 \) (constant), the steady-mode logit simplifies to \( a^{(h)} = \phi^{(h)} / \omega_\tau^{(h)} = \phi^{(h)} \). Thus, \(a^{(h)} \approx \sigma\big(w^{(h)} x + b^{(h)}\big) \) for chosen weights \( w^{(h)}, b^{(h)} \), emulating a sigmoid hidden unit.\\
\textbf{Attention Weights and output:} For \( T=1 \), the softmax over one ``key'' yields \( \alpha^{(h)} = \exp(a^{(h)}) / \exp(a^{(h)}) = 1 \). The head output is \( y^{(h)} = \int_T\alpha^{(h)} v^{(h)}dt \). Set \( v_{\text{proj}} \) such that \( v^{(h)} = 1 \) (scalar), yielding \(y^{(h)} \approx \sigma\big(w^{(h)} x + b^{(h)}\big)\). For vector-valued \( v^{(h)} \), more complex combinations are possible, but scalars suffice here.\\
\textbf{Output Projection:} Concatenate head outputs: \( Y = [y^{(1)}; y^{(2)}; \dots; y^{(H)}] \in \mathbb{R}^{H} \). Apply the final dense layer:
\begin{equation}
g(x) = (Y \cdot W_o) + b_o \in \mathbb{R}^m.
\end{equation}
With \( y^{(h)} \approx \sigma\big(w^{(h)} x + b^{(h)}\big) \), this matches a single-hidden-layer network with \( H \) units. By the UAT, for large \( H \), such networks approximate any continuous \( f \) on compact \( K \) to accuracy \( \epsilon \), by choosing appropriate \( w^{(h)}, b^{(h)}, W_o, b_o \).

\section{Training, Gradients and Complexity}

\subsection{Gradient Characterization}\label{appendix:sensitivty_analysis}

We analyze the sensitivity of the dynamics with respect to the underlying learnable parameters. Specifically, we compute closed-form derivatives of both the steady state and the full trajectory \(a_t\) with respect to the parameters \(\phi\) and \(\omega_\tau\). These expressions illuminate how gradients flow through the system, and provide guidance for selecting parameterizations that avoid vanishing or exploding gradients.

\subsubsection{Trajectory sensitivities for Closed-form formulation}
The trajectory is given by
\begin{equation}
a_t = a^* + (a_0-a^*) e^{-\omega_\tau t},
\end{equation}
which depends on \((\phi, \omega_\tau)\) both through the equilibrium \(a^*\) and the exponential term.\\
\textbf{Derivative with respect to \(\phi\):} We obtain
\begin{equation}
\frac{\partial a_t}{\partial \phi} = \frac{1 - e^{-\omega_\tau t}}{\omega_\tau}
\end{equation}
\emph{Interpretation}: For large \(\omega_\tau\), the gradient with respect to \(\phi\) saturates quickly but shrinks to scale \(\mathcal{O}(1/\omega_\tau)\), potentially slowing learning of \(\phi\). Conversely, very small \(\omega_\tau\) leads to large steady-state gradients, which may destabilize optimization.\\
\textbf{Derivative with respect to \(\omega_\tau\):} Here, both the equilibrium and the decay rate depend on \(\omega_\tau\), yielding
\begin{equation}
 \frac{\partial a_t}{\partial \omega_\tau}
= -\frac{\phi}{\omega_\tau^2}\big(1 - e^{-\omega_\tau t}\big)
- (a_0-a^*)\,t\, e^{-\omega_\tau t}.   
\end{equation}
\emph{Interpretation}: The gradient with respect to \(\omega_\tau\) contains a transient term proportional to \(t e^{-\omega_\tau t}\), which dominates at intermediate times, and a steady-state contribution proportional to \(-\phi/\omega_\tau^2\), which persists asymptotically. Thus, sensitivity to \(\omega_\tau\) is time-dependent, peaking before vanishing exponentially in the transient component.

\subsection{Gradient-Based Training}\label{appendix:training}
Like Neural ODEs \cite{chen2018neural} and CT-RNNs \cite{rubanova2019latent}, NAC produces differentiable computational graphs and can be trained using gradient-based optimization, such as the adjoint sensitivity method~\cite{cao2003adjoint} or backpropagation through time (BPTT)~\cite{lecun1988theoretical}. In this work, we use BPTT exclusively, as the adjoint sensitivity method can introduce numerical errors~\cite{zhuang2020adaptive}.

\section{Evaluation} \label{appendix:evaluation}

\begin{table*}[h!]
\centering
\caption{Summary of Key Hyperparameters of All Experiments}
\label{tab:hparams}
\resizebox{0.9\textwidth}{!}{%
\begin{tabular}{lcccccc}
\toprule
\textbf{Param.} & \textbf{E-MNIST} & \textbf{PAR} & \textbf{LRD} & \textbf{CarRacing} & \textbf{Udacity} & \textbf{RUL EST.} \\
\midrule
Conv layers & 2×\textbf{1D}(64@5) & \textbf{1D}(64@5, 64@3) & -- & 3×TD-\textbf{2D}(10--30@3--5) & 5×\textbf{2D}(24--64@5--3, ELU) & 2×\textbf{1D}(32@3, 16@2) \\
NAC & 64-d, 8h & 32-d, 4h & 64-d, 16h  & 64-d, 16h & 100-d, 20h & 16-d, 8h \\
Seq\_len & 256 & 1 & 96  & 1 & 1 & 1000 \\

Dense & 32--10(SM) & 32--11(SM) & -- & 64--5(SM) & 64--1(Lin) & 1(Lin) \\
Dropout & \textendash & \textendash & \textendash & 0.2 & 0.5 & \textendash \\
Opt. & AdamW & AdamW & AdamW & Adam & AdamW & AdamW \\
LR & 0.001 & 0.001  0.001 & 0.0001 & 0.001 & 0.001 \\
Loss & SCE & SCE & MSE & SCE & MSE & MSE \\
Metric & Acc & Acc & MSE & Acc & MSE & Score \\
Batch & 32 & 20 & 64 & 32 & 40 & 32 \\
Epochs & 150 & 100 & 50 & 100 & 10 & 150 \\
\bottomrule
\end{tabular}}
\begin{minipage}{\textwidth}
\footnotesize
\textbf{Note:} SCE = Sparse Categorical Crossentropy; Acc = Accuracy; MAE = Mean Absolute Error; MSE = Mean Squared Error; SM = softmax; Lin = Linear; TD = TimeDistributed; Conv1D/2D = Conv1D/2D; $d$ = model dimension; $h$ = attention heads.\\[1pt]
\vspace{-5mm}
\end{minipage}
\end{table*}

\subsection{Hyperparameter Tuning}\label{appendix:BO}
To reduce manual tuning and ensure well-chosen hyperparameters, we employ Bayesian optimization to tune all the hyperparameters. We optimize all key hyperparameters of the neural network, including the number of Conv1D layers, filters, kernel sizes, activation functions, NAC's $d_{\text{model}}$, num$\_$heads, Top-\emph{K}, sparsity. We also tune the Dense layer widths, optimizer choice, training batch size, and learning rate. The optimization is performed over 200 trials using \textit{MSE} as the training objective. The best performing hyperparameter for all experiments are summarized in Table~\ref{tab:hparams}. The brief overview of Bayesian Optimization is provided below.
\subsubsection{Overview of Bayesian Optimization}
Bayesian Optimization (BO) is a sample-efficient strategy for hyperparameter tuning of neural networks, particularly useful when the objective function \( f(\mathbf{x}) \) is expensive to evaluate and gradients are unavailable. BO builds a probabilistic surrogate model using a Gaussian Process (GP), which provides a posterior mean \( \mu(\mathbf{x}) \) and uncertainty \( \sigma(\mathbf{x}) \) over the objective function values. To select promising hyperparameter configurations \( \mathbf{x} \), BO maximizes an acquisition function, such as Expected Improvement (EI), defined by
\begin{equation}
\text{EI}(\mathbf{x}) =
\left(f(\mathbf{x}^+) - \mu(\mathbf{x}) - \xi\right)\Phi(Z)
+\sigma(\mathbf{x})\phi(Z),
\end{equation}
where \( f(\mathbf{x}^+) \) is the best observed function value, \( \xi \geq 0 \) controls the exploration-exploitation trade-off, 
\begin{equation}
Z=
\frac{f(\mathbf{x}^+) - \mu(\mathbf{x}) - \xi}
{\sigma(\mathbf{x})},
\end{equation}
and \( \Phi \) and \( \phi \) are the cumulative distribution function (CDF) and probability density function (PDF) of the standard normal distribution, respectively. Iteratively, BO proposes new hyperparameters by maximizing EI, evaluates these with the true objective, and updates the surrogate model until a predefined stopping criterion is met, such as iterations (N) \cite{snoek2012practical}. The full algorithm is provided in Algorithm \ref{algo:BO}.
\begin{algorithm}[ht!]
\caption{Optimization for Hyperparameter Tuning}
\begin{algorithmic}
\REQUIRE Objective function \(f(\mathbf{x})\),  Initial dataset \(X_0 = \{(\mathbf{x}_i, y_i)\}_{i=1}^n\) with \(y_i = f(\mathbf{x}_i)\), Number of iteration \(N\), Exploration parameter \(\xi \geq 0\)  
\FOR{\(t=1\) to \(N\)}
    \STATE Fit GP to \(X_{t-1}\), get \(\mu(\mathbf{x}), \sigma(\mathbf{x})\)
    \STATE Acquisition function:
    $\text{EI}(\mathbf{x}) =
    (f(\mathbf{x}^+)-\mu(\mathbf{x})-\xi)\Phi(Z)
    +\sigma(\mathbf{x})\phi(Z)$
    where
    \STATE $Z = \frac{f(\mathbf{x}^+) - \mu(\mathbf{x}) - \xi}{\sigma(\mathbf{x})}, \quad f(\mathbf{x}^+) = \min_{\mathbf{x}_i \in X_{t-1}} y_i$
    \STATE Select \(\mathbf{x}_t = \arg\max_{\mathbf{x}} \text{EI}(\mathbf{x})\)
    \STATE Evaluate \(y_t = f(\mathbf{x}_t)\)
    \STATE Update \(X_t = X_{t-1} \cup \{(\mathbf{x}_t, y_t)\}\)
\ENDFOR
\STATE \RETURN \(\mathbf{x}^+ = \arg\min_{\mathbf{x}_i \in X_N} y_i\)
\end{algorithmic}
\label{algo:BO}
\end{algorithm}

\subsection{Experimental Details} \label{appendiX:experiments}

\subsubsection{Event-based MNIST}\label{appendix:mnist}
\textbf{Dataset Explanation and Curation:} The MNIST dataset, introduced by \cite{deng2012mnist}, is a widely used benchmark for computer vision and image classification tasks. It consists of 70,000 grayscale images of handwritten digits (0–9), each of size \(28\times28\) pixels, split into 60,000 training and 10,000 testing samples.\\
\textbf{Preprocessing:} We follow the preprocessing pipeline described in \cite{lechner2022mixed}, which proceeds as follows. First, a threshold is applied to convert the 8-bit pixel values into binary values, with 128 as the threshold on a scale from 0 (minimum intensity) to 255 (maximum intensity). Second, each \(28\times28\) image is reshaped into a one-dimensional time-series of length 784. Third, the binary time-series is encoded in an event-based format, eliminating consecutive occurrences of the same value; for example, the sequence \([1,1,1,1]\) is transformed into \((1, t=4)\). This encoding introduces a temporal dimension and compresses the sequences from 784 to an average of 53 time steps. Finally, to facilitate efficient batching and training, each sequence is padded to a fixed length of 256, and the time dimension is normalized such that each symbol corresponds to one unit of time. The resulting dataset defines a per-sequence classification problem on irregularly sampled time-series. \\
\textbf{Neural Network Architecture:} We develop an end-to-end hybrid neural network by combining compact convolutional layers with NAC or counterparts baselines for fair comparison. Detailed hyperparameters and architectural specifications are provided in Table \ref{tab:hparams}.

\subsubsection{Person Activity Recognition (PAR)}\label{appendix:health}
\textbf{Dataset Explanation and Curation:} We used the Localized Person Activity Recognition dataset provided by UC Irvine \cite{localization_data_for_person_activity_196}. The dataset comprises 25 recordings of human participants performing different physical activities. The eleven possible activities are “walking,” “falling,” “lying down,” “lying,” “sitting down,” “sitting,” “standing up from lying,” “on all fours,” “sitting on the ground,” “standing up from sitting,” and “standing up from sitting on the ground.” The objective of this experiment is to recognize the participant’s activity from inertial sensors, formulating the task as a per-time-step classification problem. The input data consist of sensor readings from four inertial measurement units placed on participants’ arms and feet. While the sensors are sampled at a fixed interval of 211 ms, recordings exhibit different phase shifts and are thus treated as irregularly sampled time-series.\\
\textbf{Preprocessing:} We first separated each participant’s recordings based on sequence identity and calculated elapsed time in seconds using the sampling period. To mitigate class imbalance, we removed excess samples from overrepresented classes to match the size of the smallest class. Subsequently, the data were normalized using a standard scaler.\\
\textbf{Neural Network Architecture:} Following the approach in Section \ref{appendix:mnist}, we developed an end-to-end neural network combining convolutional heads with NAC or other baselines. Hyperparameter details are summarized in Table \ref{tab:hparams}.

\subsubsection{Multivariate Long-range Dependencies (LRDs)}\label{appendix:lrd}
\textbf{Dataset Explanation and Curation:} We utilized two well-known multivariate long-range datasets to evaluate NAC’s capabilities in modeling long-range dependencies: (i) Electrical Transformer Temperature (ETTm1) and (ii) Jena Climate. ETTm1 contains 7 features and is used for long-term electricity transformer load forecasting. It provides data at 15-minute intervals, capturing real-world temporal patterns for time-series analysis. Jena Climate contains 14 features capturing various atmospheric and environmental measurements. It provides data at 10-minute intervals for modeling and forecasting climate-related time series.\\
\textbf{Preprocessing:} We used 12 hours of data from ETTm1 to train NAC and the baseline models, and the following 6 hours were forecasted. Similarly, we used 8 hours of data from Jena Climate for training and the subsequent 4 hours for forecasting. All features were normalized using a standard scaler before training.\\
\textbf{Neural Network Architecture:} The neural network consists of a single NAC layer with $d_{\text{model}}$ = 64, 16 heads, Top-\emph{K}=32 and 50\% sparsity. Detailed hyperparameters are provided in Table~\ref{tab:hparams}.

\subsubsection{Autonomous Vehicle}\label{appendix:AVs}
\textbf{Dataset Explanation and Curation:} We followed the data collection methodology described in \cite{razzaq2023neural}. For OpenAI-CarRacing, a Proximal Policy Optimization (PPO) trained agent (5M timesteps) was used to record 20 episodes, yielding approximately 48174 RGB images of size \(92\times92\times3\) with corresponding action labels across five discrete actions (no-act, move left, forward, move right, stop). For the Udacity simulator, we manually controlled the vehicle for 50 minutes, producing 15647 RGB images of size \(320\times160\times3\), captured from three camera streams (left, center, right) along with their corresponding continuous steering values.\\
\textbf{Preprocessing:} No preprocessing was applied to the OpenAI-CarRacing dataset. For the Udacity simulator, we followed the preprocessing steps in \cite{shibuya_car_behavioral_cloning}. Each image was first cropped to remove irrelevant regions and resized to \(66\times120\times3\). Images were then converted from RGB to YUV color space to match the network input. To improve robustness, data augmentation techniques, including random flips, translations, shadow overlays, and brightness variations, were applied to simulate lateral shifts and diverse lighting conditions.\\
\textbf{Neural Network Architecture:} For OpenAI-CarRacing, we modified the neural network architecture proposed in \cite{razzaq2023neural}, which combines compact CNN layers for spatial feature extraction with LNNs to capture temporal dynamics. In our implementation, the LNN layers were replaced with NAC and its comparable alternatives for fair evaluation. Full hyperparameter configurations are provided in Table \ref{tab:hparams}. For the Udacity simulator, we modified the network proposed in \cite{bojarski2016end} by replacing three latent MLP layers with NAC and its counterparts. Full hyperparameters are summarized in Table \ref{tab:hparams}.\\

\subsubsection{Industry 4.0}\label{appendix:industry}

\textbf{Dataset Explanation and Curation:} \textit{PRONOSTIA dataset} is a widely recognized benchmark dataset in the field of condition monitoring and degradation estimation of rolling-element bearings. Nectoux et al.\cite{nectoux2012pronostia} developed this dataset as part of the PRONOSTIA experimental platform. The dataset comprises 16 complete run-to-failure experiments performed under accelerated wear conditions across three different operating settings: 1800 rpm with 4 kN radial load, 1650 rpm with a 4.2 kN load, and 1500 rpm with a 5 kN load, all at a frequency of 100 Hz. Vibration data were recorded using accelerometers placed along the horizontal and vertical axes, which were sampled at 25.6 kHz. Additionally, temperature readings were collected at a sampling rate of 10 Hz. \\
\textit{XJTU-SY Dataset} is another widely recognized benchmark dataset developed through collaboration between Xi'an Jiaotong University and Changxing Sumyoung Technology \cite{wang2018xjtu}. The dataset comprises 15 complete run-to-failure experiments performed under accelerated degradation conditions with three distinct operational settings: 1200 rpm (35 Hz) with a 12 kN radial load, 2250 rpm (37.5 Hz) with an 11 kN radial load, and 2400 rpm (40 Hz) with a 10 kN radial load. Vibrational signals were recorded using an accelerometer mounted on the horizontal and vertical axes and sampled at 25.6 kHz. This dataset is only used for the cross-validation test.\\
\textit{HUST Dataset} is a practical dataset developed by Hanoi University of Science and Technology to support research on ball bearing fault diagnosis \cite{hong2023hust}. The dataset includes vibration data collected from five bearing types (6204, 6205, 6206, 6207, and 6208) under three different load conditions: 0 W, 200 W, and 400 W. Six fault categories were introduced, consisting of single faults (inner race, outer race, and ball) and compound faults (inner–outer, inner–ball, and outer–ball). Faults were created as early-stage defects in the form of 0.2 mm micro-cracks, simulating real degradation scenarios. The vibration signals were sampled at 51.2 kHz with approximately 10-second recordings for each case. This dataset is only used for the cross-validation test.\\
\textbf{Preprocessing:} Condition monitoring data comprises 1D non-stationary vibrational signals collected from multiple sensors. To extract meaningful information, these signals must be transformed into features that possess meaningful physical interpretability. We utilized the preprocessing proposed in \cite{razzaq2025carle}, and physically plausible labels are generated according to \cite{razzaq2025developingdistanceawareuncertaintyquantification}. Initially, the signal is segmented into small, rectangularized vectors using a windowing technique, enabling better localization of transient characteristics. The continuous wavelet transform (CWT) with the Morlet wavelet as the mother wavelet is then applied to obtain a time-frequency representation (TFR). The CWT is defined as \(\Gamma(a, b) = \int_{-\infty}^{\infty} x_w(t) \frac{1}{\sqrt{a}} \psi^*\left( \frac{t - b}{a} \right) \, dt\), where \( a \) and \( b \) denote the scale and translation parameters, respectively, and \( \psi \) is the Morlet wavelet function. From the TFR, a compact set of statistical and domain-specific features is extracted to characterize the operational condition of the bearing.\\
\textbf{Neural Network Architecture:} The objective of this problem is to design a compact neural network that can effectively model degradation dynamics while remaining feasible for deployment on resource-constrained devices, enabling localized and personalized prognostics for individual machines. To achieve this, we combine a compact convolutional network with NAC. The CNN component extracts spatial degradation features from the training data, while NAC performs temporal filtering to emphasize informative features. This architecture maintains a small model size without sacrificing representational capacity. Full hyperparameter configurations are reported in Table \ref{tab:hparams}.\\
\textbf{Evaluation Metric:} Score is a metric specifically designed for RUL estimation in the IEEE PHM \cite{nectoux2012pronostia} to score the estimates. The scoring function is asymmetric, penalizing overestimations more heavily than early predictions. This reflects practical considerations, as late maintenance prediction can lead to unexpected failures with more severe consequences than early intervention can.
\begin{equation}
\textit{Score} = \sum_{i : \hat{y}_i < y_i} \left( e^{-\frac{\hat{y}_i - y_i}{13}} - 1 \right) + \sum_{i : \hat{y}_i \geq y_i} \left( e^{\frac{\hat{y}_i - y_i}{10}} - 1 \right)
\label{eq:score}
\end{equation}

\end{document}